\pdfoutput=1
\documentclass[11pt]{article}

\usepackage[]{acl}

\usepackage{times}
\usepackage{latexsym}
\usepackage[T1]{fontenc}
\usepackage[utf8]{inputenc}
\usepackage{microtype}

\usepackage{booktabs} 

\usepackage{xcolor}
\usepackage{hyperref}
\usepackage{graphicx}
\usepackage{adjustbox}
\usepackage{amssymb}
\usepackage{pifont}
\usepackage{multirow}
\usepackage{amsmath}
\usepackage{bbm}
\usepackage{float}
\usepackage{caption}
\usepackage{subcaption}

\usepackage{colortbl}

\usepackage[
  separate-uncertainty = true,
  multi-part-units = repeat
]{siunitx}

\title{Relational Sentence Embedding for Flexible Semantic Matching}

\author{
Bin Wang\textsuperscript{\rm \dag},~~~~~~Haizhou Li\textsuperscript{\rm $\natural$,\rm $\S$,\rm \dag}
\\
\textsuperscript{\rm \dag}National University of Singapore, Singapore
\\
\textsuperscript{\rm $\natural$}School of Data Science, The Chinese University of Hong Kong, Shenzhen, China
\\
\textsuperscript{\rm $\S$}Shenzhen Research Institute of Big Data
\\
\texttt{bwang28c@gmail.com}} 

\begin{document}
\maketitle
\begin{abstract}

    We present Relational Sentence Embedding (RSE), a new paradigm to further discover the potential of sentence embeddings. Prior work mainly models the similarity between sentences based on their embedding distance. Because of the complex semantic meanings conveyed, sentence pairs can have various relation types, including but not limited to entailment, paraphrasing, and question-answer. It poses challenges to existing embedding methods to capture such relational information. We handle the problem by learning associated relational embeddings. Specifically, a relation-wise translation operation is applied to the source sentence to infer the corresponding target sentence with a pre-trained Siamese-based encoder. Later, the fine-grained relational similarity scores can be estimated with learned relational embeddings. We benchmark our method on 19 datasets covering a wide range of tasks, including semantic textual similarity, transfer, and domain-specific tasks. Experimental results show that our method is effective and flexible in modelling sentence relations and outperforms a series of state-of-the-art sentence embedding methods.\footnote{Our code is available \url{https://github.com/BinWang28/RSE}}


\end{abstract}

\section{Introduction}

    Sentence representation learning is a long-standing topic in natural language understanding and has broad applications in many tasks, including retrieval, clustering, and classification \cite{cer-etal-2018-universal}. It has been studied extensively in existing literature \cite{kiros2015skip,conneau-etal-2017-supervised,reimers-gurevych-2019-sentence,gao2021simcse} and the problem is defined as learning vector representation ($e_i$) for sentence ($s_i$) and the similarity between sentences ($s_i$,~$s_j$) is inferred by their pairwise cosine similarity: $cos(e_i,e_j)$. Prior work found that the entailment relation in natural language inference (NLI) data aligns well with the semantic similarity concept defined in STS datasets \cite{conneau-etal-2017-supervised,reimers-gurevych-2019-sentence,gao2021simcse}. However, the concept of similarity is vaguely defined, and the relationship between sentences is indeed multifaceted \cite{wang2022just}. One can only rely on experimental verification to determine what relation between sentences aligns well with human perception of ``semantic similarity''. The similarity intensity between sentences varies greatly based on the perspective of personal understanding. 
    
    \begin{table}[t]
        \centering
        \begin{adjustbox}{width=0.45\textwidth,center}
        \begin{tabular}{ l  c  c  c  c  c  c  c c }
        \toprule
         \textbf{Dataset} & \textbf{Relation} & \textbf{STS\_avg} \\
        \midrule
        SNLI+MNLI & \textsuperscript{*}Entailment & 81.33 \\ \midrule
        SNLI+MNLI & Entailment & 77.67 \\
        QQP (149K) & Duplicate Question & 75.96 \\
        Flicker (318K) & Same Caption & 74.55 \\
        ParaNMT (5M) & Paraphrase  & 75.32 \\
        QNLI (55K) & Question Answer  & 72.21 \\
        \midrule 
        All-Above & All 5 relations & 74.62 \\
        \midrule \midrule
        \textbf{RSE} & \textbf{\emph{Mixed}} & \textbf{82.17} \\ \midrule
        WIKI (1M) & Dropout & 75.45 \\
        \bottomrule
        \end{tabular}
        \end{adjustbox}
        \caption{Comparison of different supervised signals as positive sentence pairs. Averaged results on STS datasets are reported with \emph{BERT\textsubscript{base}} model. \textsuperscript{*} indicates the model is trained with hard negatives using contradiction pairs \cite{gao2021simcse}. Experimental details can be found in Appendix \ref{appendix:mscs}.}
        \label{tab:demo1}
    \end{table}

    \begin{figure*}[t]
        \centering
        
        \begin{subfigure}[b]{0.26\textwidth}
        \centering
         \includegraphics[width=\textwidth]{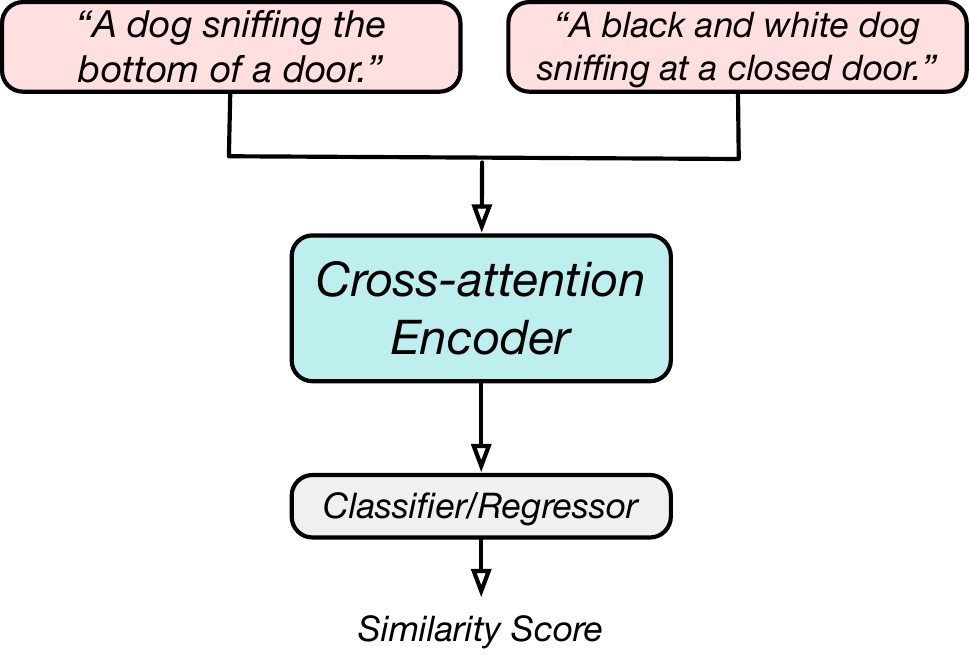}
        \caption{Cross-Attention Matching}
        \end{subfigure}
        \hfill
        \begin{subfigure}[b]{0.26\textwidth}
        \centering
         \includegraphics[width=\textwidth]{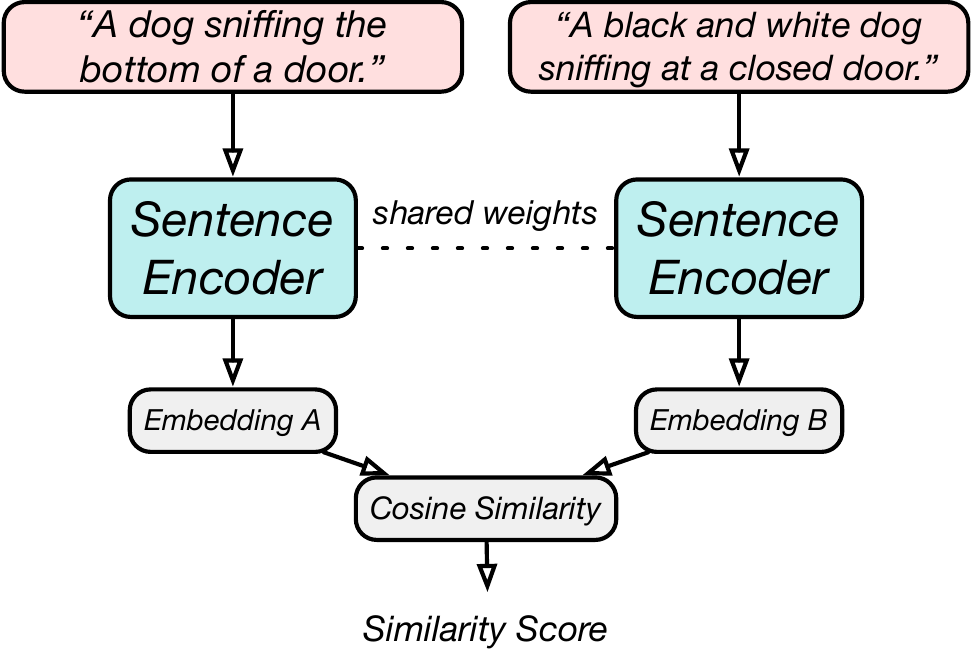}
        \caption{Sentence Encoder Matching}
        \label{fig:sent-rank-b}
        \end{subfigure}
        \hfill
        \begin{subfigure}[b]{0.4\textwidth}
        \centering
         \includegraphics[width=\textwidth]{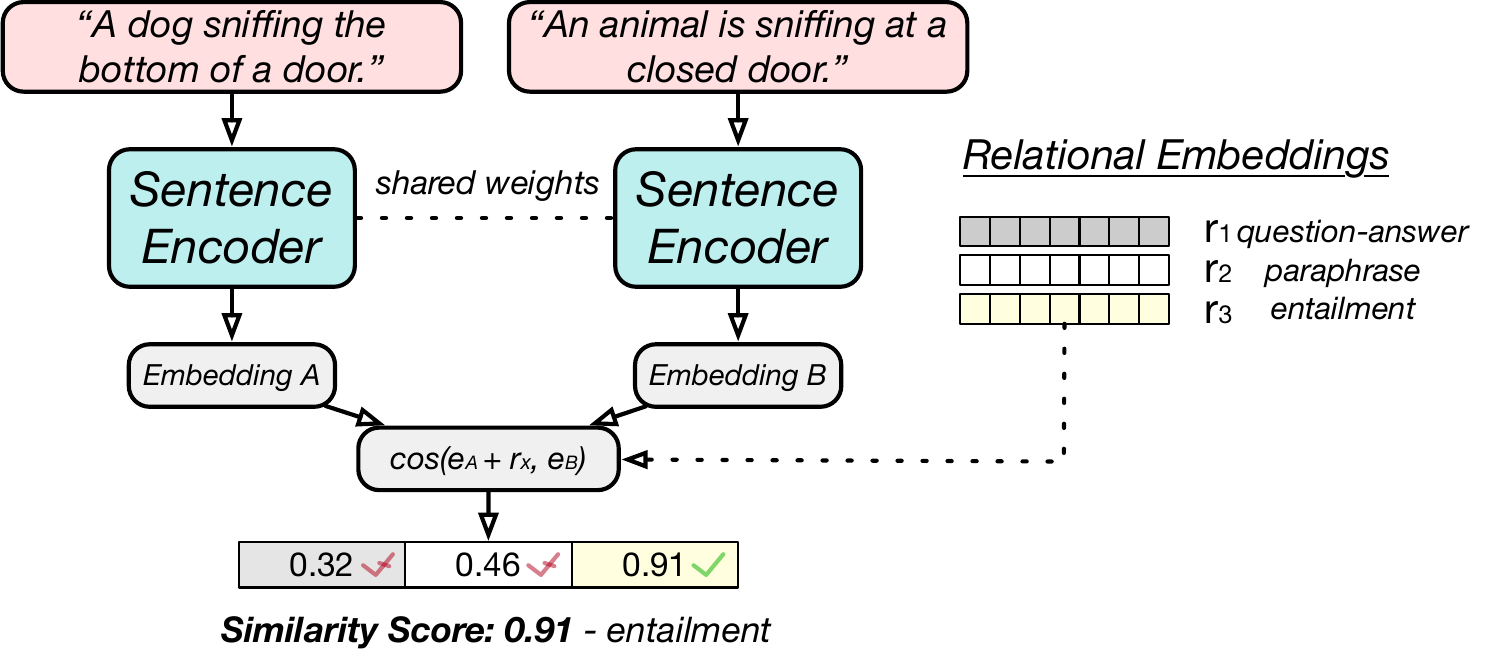}
        \caption{Relational Sentence Embedding}
        \label{fig:sent-rank-c}
        \end{subfigure}
        \caption{Comparison of different sentence-pair matching architectures. Relational sentence embedding adopts relation-specific embeddings for more flexible semantic matching with interpretable results. Meantime, the Siamese network allows much faster inference speech compared with cross-attention matching.}
        \label{fig:sent-rank}
    \end{figure*}

   In this work, we do not focus on discussing the detailed definition of semantic relations but focus on the explicit modelling of sentence relations. Instead of treating sentence pairs as either positive or negative samples in contrastive learning \cite{gao2021simcse,yan2021consert,wu2021esimcse}, we treat sentence pairs as triples: ($s_i$, \emph{relationship}, $s_j$), with relational information embedded. We learn a translational operation between sentence embeddings to contrastively model the relationship between sentences. Despite its simplicity, the relational modelling component largely expands the concept of existing sentence embedding and introduces unique properties for more flexible semantic matching capabilities. Our method is called \emph{Relational Sentence Embedding} (RSE).

    

    
    
    First, with relation modelling, RSE can leverage multi-source relational data for improved generalizability. Earlier work mainly focuses on using NLI data for supervised sentence embedding learning, as the best performance is witnessed. It remains an under-explored problem to learn from multi-source relational data jointly for sentence representation learning. Table~\ref{tab:demo1} shows our experimental study to combine multi-source relational data as the positive samples for contrastive sentence embedding learning \cite{gao2021simcse}. Simply merging all relational data leads to performance down-gradation (ref.~'All-Above'). In contrast, with explicit relation modelling by translation operations, RSE can effectively leverage multi-source relational information into a shared sentence encoder while providing exceptional performance.
    
    Second, unlike conventional SE, RSE can provide fine-grained relational similarity scores of sentence pairs. As shown in Figure~\ref{fig:sent-rank-b}, conventional sentence embedding computes one unified similarity score for sentence pairs. It is less flexible and interpretable as no detailed information is provided about what aspects of sentences are similar. In contrast, as shown in Figure~\ref{fig:sent-rank-c}, RSE can infer fine-grained sentence similarity scores according to the relation of interest, which are critical features for heterogeneous applications (e.g., retrieval and clustering). RSE can vastly extend the sentence embedding concept and has excellent potential in methodology development and applications.
    
    We benchmark our approach on 19 datasets to verify the effectiveness and generalizability of our approach. For semantic textual similarity, we conduct experiment on seven standard STS \cite{agirre-etal-2012-semeval,agirre-etal-2013-sem,agirre-etal-2014-semeval,agirre-etal-2015-semeval,agirre-etal-2016-semeval,cer-etal-2017-semeval,marelli-etal-2014-sick} datasets and STR dataset \cite{abdalla2021makes}. RSE achieves an averaged Spearsman's correlation of 82.18\% with BERT\textsubscript{\texttt{base}} model on STS datasets. We also achieve SOTA performance on 7 transfer tasks \cite{conneau-kiela-2018-senteval}, and 4 domain-specific tasks \cite{wang-etal-2021-tsdae-using}. We also present the case study and in-depth discussions.

\section{Related Work}

    Sentence embedding can be divided into unsupervised and supervised approaches. Even though unsupervised methods have greatly improved with effective data augmentation techniques for contrastive learning \cite{gao2021simcse,wu2021esimcse,chen2022generate}, supervised methods still outperform their unsupervised counterparts. The success of supervised sentence embedding is mainly because of the availability of high-quality natural language inference (NLI) data \cite{bowman-etal-2015-large,williams-etal-2018-broad}. InferSent \cite{conneau-etal-2017-supervised}, SBERT \cite{reimers-gurevych-2019-sentence}, and SimCSE \cite{gao2021simcse} are three milestones on sentence embedding that effectively leverage NLI data using LSTM, Pre-trained BERT and contrastive learning, respectively. The NLI data covers three sentence relations: entailment, contradiction and neutral. The relationships between sentences are manifold. However, prior work found that training with other supervised signals \cite{young-etal-2014-image,wieting2017paranmt} does not achieve comparable performance as NLI data \cite{gao2021simcse}. In this work, we focus on learning from multi-source relational signals for better sentence representation learning.

    \begin{figure*}[t]
        \centering
         \includegraphics[width=0.7\textwidth]{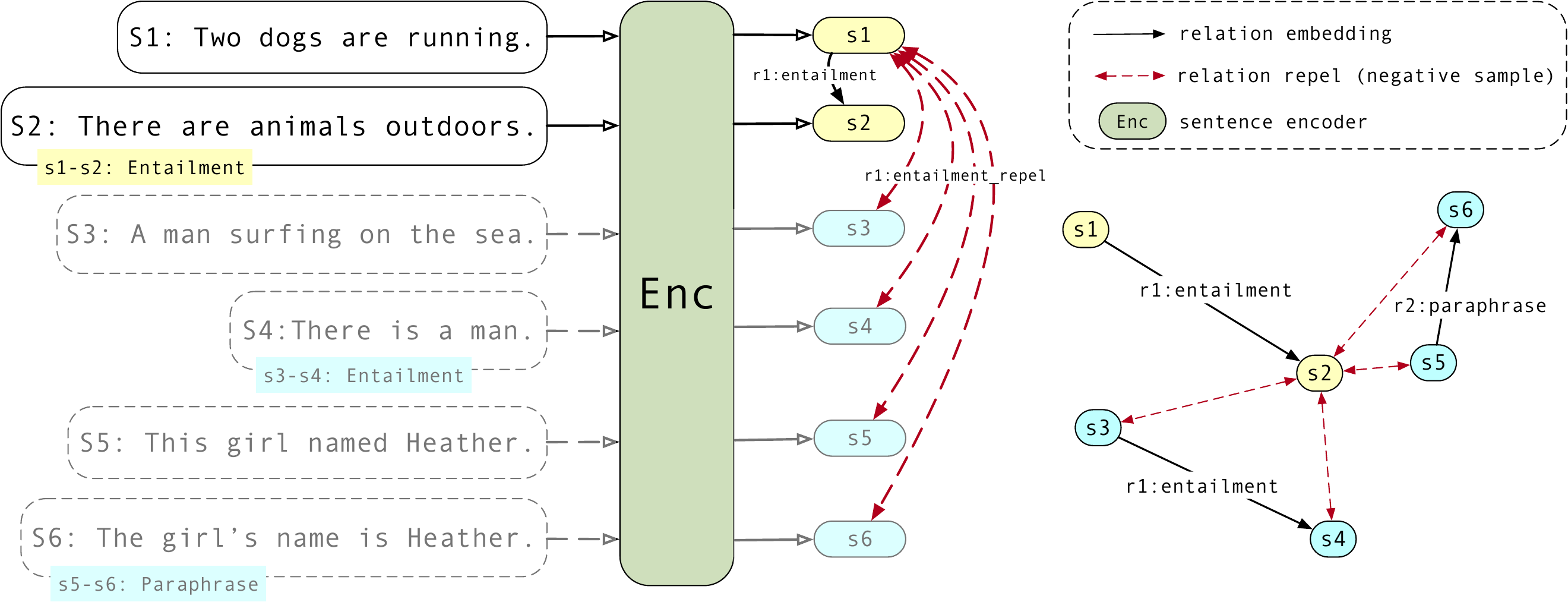}
        \caption{RSE takes the corresponding relation transformation as positive triples and treats other in-batch samples as negatives.}
        \label{fig:RSE-framework}
    \end{figure*}
    
    Knowledge graph embeddings are designed for relational data modelling on triples: $(h,r,t)$. The methods can be classified into structural-based models and semantic-enhanced models. The structural-based methods learn relational embedding and entity embedding jointly through scoring functions like TransE \cite{bordes2013translating}, DistMult \cite{yang2014embedding} and RotatE \cite{sun2019rotate}. Semantic-enhanced models incorporate additional information for entities, including entity types and textual descriptions, when learning embeddings. TEKE \cite{wang2016text}, InductivE \cite{wang2021inductive} and KG-BERT \cite{yao2019kg} belong to this type. For sentence representation learning from multi-source supervised signals, the sentence pairs can be viewed as triples: $(s_i, relationship, s_j)$. Therefore, we optimize the semantic representation of sentences and perform translation operations between them to model their respective relations. An analogy to the word-level relational graph as Word-Net \cite{miller1995wordnet}, and Relational Word Embedding \cite{camacho-collados-etal-2019-relational}, our work can also be viewed as a pioneering study on the sentence level.

\section{Relational Sentence Embedding}

        
        Given $n$ pre-defined relations $R=\{r_1, r_2, ..., r_n\}$, the multi-source relation sentence pair datasets are composed of triples: $D=\{(s_i, r_k, s_j)\}$, where $s_i$ and $s_j$ corresponds to the $i_{th}$ and $j_{th}$ sentence. The goal of relation sentence embedding is to learn sentence representations and relational embeddings jointly. The relational similarity intensity between sentences can be inferred from learned embeddings through either simple scoring functions or complex transformations. In this work, as a prototype of RSE, we use translation operation \cite{bordes2013translating} for sentence relation modelling.

        
    

        

    \subsection{Relational Contrastive Learning}
    
        Contrastive learning approaches have demonstrated their effectiveness in sentence embedding learning \cite{gao2021simcse,janson2021semantic}. The core concept is learning representations that pull together close concepts and push away non-related ones. To build contrastive samples, we leverage the relational signal from sentence triples of supervised data. Assume we have relational triples $(s_i, r_k, s_j)$, let $\mathbf{h_i}$ and $\mathbf{h_j}$ denote the representation of $s_i$ and $s_j$ obtained from sentence encoder $\mathbf{h_i} = f_\theta(s_i)$ and $\mathbf{h^r_k}$ denote the relational embedding of the $k_{th}$ relation. We then treat $\mathbf{h_i}+\mathbf{h^r_k}$ and $\mathbf{h_j}$ as positive pair. Following the previous contrastive learning approach~\cite{chen2020simple}, other in-batch tail sentences are treated as negatives. Assuming we have $N$ sentence triples in one mini-batch, the training objective can be written as follow:
        
        \begin{equation}
            \mathcal{L}_i = - \log \frac{ e^{sim(\mathbf{h_i+h_k^r},\mathbf{h_j})/ \tau} }{\sum_{m=1}^N e^{sim(\mathbf{h_i+h_k^r},\mathbf{h_m})/ \tau} }
        \label{eqa:rcs}
        \end{equation}
        where $sim(\cdot,\cdot)$ is the cosine similarity and $\tau$ is the temperature hyperparameter. The pre-trained BERT \cite{devlin-etal-2019-bert} or RoBERTa \cite{liu2019roberta} model $f_\theta(\cdot)$ is used to obtain sentence embedding. The relation embedding $\mathbf{h^r_k}$ is randomly initialized and jointly optimized with the pre-trained language model with the above relational contrastive objective.
        
        Figure~\ref{fig:RSE-framework} is a graphical illustration of Eq.~\ref{eqa:rcs}. The relational information is learned by relational embedding through translation operation. That is, for triple $(s_i, r_k, s_j)$, we expect $\mathbf{h_i}+\mathbf{h_k^r}\approx \mathbf{h_j}$. The embedding property can further infer the relation intensity between sentences during inference.
        
        In knowledge graph embedding, there are more diverse ways to model relations, including simple scoring functions (TransE~\cite{bordes2013translating}, TransR~\cite{lin2015learning} and RotatE~\cite{sun2019rotate}), semantic matching methods (DistMult~\cite{yang2014embedding}, TUCKER \cite{balazevic-etal-2019-tucker}) and neural network methods (ConvE~\cite{dettmers2018convolutional}). In this work, we experiment with the simple translational operation for relation modeling for simplicity and interpretability. Other complex relation modeling methods can be explored in future work as they are more expressive in certain relation types, including symmetry, anti-symmetry, inversion, and composition~\cite{sun2019rotate}.


        
        
        

        


        
        
    \subsection{Supervised Signals}
    \label{subsec:supervised-signal}
    
        The key to contrastive sentence embedding is building positive and negative samples. ConSERT~\cite{yan2021consert} proposed to use adversarial attach, token shuffling, feature cutoff, and dropout as positive pair augmentation for unsupervised sentence embedding. SimCSE~\cite{gao2021simcse} further explores the potential of dropout as positive-view augmentation. In this work, we collect multi-source relational data as positive samples and view the rest in-batch samples as negatives. In general, multi-source relational data can be either supervised or unsupervised. For unsupervised learning, we can view each data augmentation method as one relation and jointly learn from multiple data augmentation methods. In this work, we focus on artificial sentence triples and explore how to combine multiple human-labelled datasets with different relation types.
        
        We deploy the following datasets for sentence triples ($s_i, r_k, s_j$). 
        1)~SNLI~\cite{bowman-etal-2015-large} + MNLI~\cite{williams-etal-2018-broad}: contain natural language inference data. The relations are \emph{entailment} ($r_1$), \emph{contradiction} and \emph{neutral}. We use \emph{entailment} relation with \emph{contradiction} relation as hard negatives. 
        2)~QQP\footnote{\url{https://data.quora.com/First-Quora-Dataset-Release-Question-Pairs}}: The Quora Question dataset. The label for each question pair is either \emph{duplication-question} ($r_2$) or \emph{non-duplicate-question}. We include the triples with \emph{duplication-question} relation.
        3)~ParaNMT~\cite{wieting2017paranmt}: \emph{paraphrase} ($r_3$) dataset generated by back-translation with filtering. 
        4)~Flicker~\cite{young-etal-2014-image}: image captioning dataset. Each image is with 5 human-written captions. We treat the captions for the same image as relation \emph{same-caption} ($r_4$). 
        5)~QNLI~\cite{wang-etal-2018-glue}: contains question-answer pairs. The label is \emph{qa-entailment} ($r_5$) or \emph{qa-not-entailment} based on whether the sentences contain the true answer to the question. We only use \emph{qa-entailment} triples.
        6)~Dropout~\cite{gao2021simcse}: is constructed in an unsupervised way. We use the Wiki-1M dataset and treat the sentence with different random dropouts as one relation. We call this relation \emph{same-sent} ($r_6$).
        We leverage all samples from the NLI dataset and at most 150K samples for other relations as the size of QQP for easy comparison.

    \subsection{Mining Hard Negatives}
    
        Contrastive learning commonly uses in-batch sentences as negative samples \cite{janson2021semantic}. However, the in-batch negatives are limited in relational sentence embeddings. It is because the in-batch samples are coming from different relation triples. The sentences from other relations are less likely to be confused with the current anchor triple. Therefore, we extend the $(s_i, r_k, s_j)$ triple to $(s_i, r_k, s_j, s_j^{-})$ with an extra in-relation negative sample $s_j^{-}$. The $s_j^{-}$ is one random sampled sentence from the triple of relation $r_k$. To better understand, the in-relation negative for relation \emph{duplicate-question} will be a random question, a `harder' negative than in-batch sentences. For the \emph{entailment} relation, we use \emph{contradiction} sentence as the hard negative. With the hard negative, the training objective can be written as follows (Ref. to ~Eq.\ref{eqa:rcs}):
        \begin{equation}
            -\log\!\frac{ e^{sim(\mathbf{h_i+h_k^r},\mathbf{h_j})/\!\tau} }{\sum\limits_{m=1}^N\!(e^{sim(\mathbf{h_i+h_k^r},\mathbf{h_m})/\!\tau}+e^{sim(\mathbf{h_i+h_k^r},\mathbf{h_m^{-}})/\!\tau})}
        \end{equation}

        As multiple relation triples are mixed in each batch during training, the batch size for each relation becomes smaller. Contrastive learning favors a relatively large batch size to reach good performance. \citet{wu2021esimcse} incorporate momentum contrast \cite{he2020momentum} for better contrastive learning. However, we did not find its contribution to the supervised setting. As the GPU memory size constrains the batch size, we scale up the batch size of contrastive learning by decoupling back-propagation between contrastive learning and decoder using Gradient Cache~\cite{gao-etal-2021-scaling}. As a result, we can train BERT\texttt{base} model with batch size 512 within an 11G GPU memory limit.
        
        
    
        
        
        
        


    
        

    
        
        

        
        


        
        

    \begin{table*}[thb]
        \centering
        \begin{adjustbox}{width=1.0\textwidth,center}
        \begin{tabular}{ l  c  c  c  c  c  c  c | c }
        \toprule
        \textbf{Model} & \textbf{STS12} & \textbf{STS13} & \textbf{STS14} & \textbf{STS15} & \textbf{STS16} & \textbf{STS-B} & \textbf{SICK-R} & \textbf{Avg.} \\
        \midrule \midrule
        \multicolumn{9}{c}{\textit{Unsupervised methods}} \\
        \midrule
        GloVe (avg.) & 55.14 & 70.66 & 59.73 & 68.25 & 63.66 & 58.02 & 53.76 & 61.32 \\
        BERT\textsubscript{\texttt{base}} (first-last-avg) & 39.70 & 59.38 & 49.67 & 66.03 & 66.19 & 53.87 & 62.06 & 56.70 \\
        CT - BERT\textsubscript{\texttt{base}} & 61.63 & 76.80 & 68.47 & 77.50 & 76.48 & 74.31 & 69.19 & 72.05 \\
        SimCSE - BERT\textsubscript{\texttt{base}} & 68.40 & 82.41 & 74.38 & 80.91 & 78.56 & 76.85 & 72.23 & 76.25 \\
        SimCSE - RoBERTa\textsubscript{\texttt{base}} & 70.16 & 81.77 & 73.24 & 81.36 & 80.65 & 80.22 & 68.56 & 76.57 \\
        \midrule
        \multicolumn{9}{c}{\textit{Supervised methods}} \\
        \midrule
        InferSent - GloVe & 52.86 & 66.75 & 62.15 & 72.77 & 66.87 & 68.03 & 65.65 & 65.01 \\
        USE & 64.49 & 67.80 & 64.61 & 76.83 & 73.18 & 74.92 & 76.69 & 71.22 \\
        SBERT\textsubscript{\texttt{base}} & 70.97 & 76.53 & 73.19 & 79.09 & 74.30 & 77.03 & 72.91 & 74.89 \\
        SimCSE - BERT\textsubscript{\texttt{base}} & 75.30 & \textbf{84.67} & 80.19 & 85.40 & 80.82 & 84.25 & 80.39 & 81.57 \\
        \midrule
        RSE - BERT\textsubscript{\texttt{base}} & \textbf{76.27}$_{\pm.34}$ & 84.43$_{\pm.23}$ & \textbf{80.60}$_{\pm.12}$ & \textbf{86.03}$_{\pm.09}$ & \textbf{81.86}$_{\pm.12}$ & \textbf{84.34}$_{\pm.13}$ & \textbf{81.73}$_{\pm.07}$ & \textbf{82.18}$_{\pm.06}$ \\ 
        RSE - BERT\textsubscript{\texttt{large}} & 76.94$_{\pm.40}$ & 86.63$_{\pm.43}$ & 81.81$_{\pm.28}$ & 87.02$_{\pm.22}$ & 83.25$_{\pm.22}$ & 85.63$_{\pm.23}$ & 82.68$_{\pm.24}$ & 83.42$_{\pm.19}$ \\ 
        \midrule
        RSE - RoBERTa\textsubscript{\texttt{base}} & 76.71$_{\pm.47}$ & 84.38$_{\pm.34}$ & 80.87$_{\pm.56}$ & 87.12$_{\pm.42}$ & 83.33$_{\pm.19}$ & 84.62$_{\pm.18}$ & 81.97$_{\pm.32}$ & 82.71$_{\pm.20}$ \\
        RSE - RoBERTa\textsubscript{\texttt{large}} & \textbf{77.81}$_{\pm.37}$ & \textbf{87.04}$_{\pm.20}$ & \textbf{82.73}$_{\pm.22}$ & \textbf{87.92}$_{\pm.25}$ & \textbf{85.44}$_{\pm.15}$ & \textbf{86.79}$_{\pm.08}$ & \textbf{83.13}$_{\pm.07}$ & \textbf{84.41}$_{\pm.08}$ \\
        \bottomrule
        \end{tabular}
        \end{adjustbox}
        \caption{Performance on STS tasks. Spearman's correlation is reported. For RSE approach, we also report the standard derivation of performance based on five runs with different random seeds. The best performance among all models and the best performance with BERT\textsubscript{\texttt{base}} backbone is displayed in \textbf{bold}.}
        \label{tab:1}
    \end{table*}

    \begin{table}[thb]
        \centering
        \begin{adjustbox}{width=0.42\textwidth,center}
        \begin{tabular}{ l  c  c  c  c  c  c  c c }
        \toprule
        \textbf{Model} & \textbf{STR} \\
        \midrule
        Unsup. SimCSE - BERT\textsubscript{\texttt{base}} & 73.98 \\
        Sup. SimCSE - BERT\textsubscript{\texttt{base}} & 80.72 \\
        \midrule
        RSE - BERT\textsubscript{\texttt{base}} & \textbf{81.23}$_{\pm.13}$ \\
        RSE - BERT\textsubscript{\texttt{large}} & 81.98$_{\pm.18}$ \\
        RSE - RoBERTa\textsubscript{\texttt{base}} & 82.57$_{\pm.30}$ \\
        RSE - RoBERTa\textsubscript{\texttt{large}} & \textbf{84.10}$_{\pm.18}$ \\
        \bottomrule
        \end{tabular}
        \end{adjustbox}
        \caption{Comparison of SimCSE and RSE performance on STR dataset.}
        \label{tab:3}
    \end{table}

        \noindent\textbf{How can we use?} As the relational information is explicitly embedded in RSE, it becomes very flexible for different applications. Given two sentence ($s_i, s_j$), we can infer their similarity score under relation $r_k$ by $f(s_i, s_j, r_k) = sim(\mathbf{h_i}+\mathbf{h^r_k},\mathbf{h_j})$. For retrieval and semantic similarity, we can use any relational similarity score or weighted sum of different relational scores as the final measure. We apply this in STS and USEB benchmarking tasks. As an additional classifier is trained, we use sentence embedding as the input for transfer tasks. We expect that with relational modelling capability, the result sentence embedding carries richer information.

\section{Experiments on STS Tasks}

    We conduct experiment on 7 semantic textual similarity datasets including STS 12-16 \cite{agirre-etal-2012-semeval,agirre-etal-2013-sem,agirre-etal-2014-semeval,agirre-etal-2015-semeval,agirre-etal-2016-semeval}, STS-Benchmark~\cite{cer-etal-2017-semeval} and SICK-Relatedness~\cite{marelli-etal-2014-sick}. We also experiment with STR-2022 \cite{abdalla2021makes}. A new STS dataset with improved label quality using comparative ranking. For a fair comparison, we report the result with the fully unsupervised setting, Spearman's correlation and `all' for result aggregation following prior works \cite{reimers-gurevych-2019-sentence,gao2021simcse}.

    \subsection{Settings}
    
        We initialize our training checkpoint with pre-trained BERT\textsubscript{\texttt{base,large}} \cite{devlin-etal-2019-bert} and RoBERTa\textsubscript{\texttt{base,large}} \cite{liu2019roberta}. The \texttt{[CLS]} token representation is taken as sentence representation ($\mathbf{h_i}$). We compare with both unsupervised and supervised sentence embedding methods, including 1) average GloVe embedding \cite{pennington-etal-2014-glove}, 2) average BERT embedding \cite{su2021whitening}, 3) CT-BERT \cite{janson2021semantic}, 4) InferSent \cite{conneau-etal-2017-supervised}, 5) USE \cite{cer-etal-2018-universal}, 6) SBERT \cite{reimers-gurevych-2019-sentence} and 7) SimCSE \cite{gao2021simcse}. For a fair comparison, we did not include T5-based methods to avoid the influence of pre-trained checkpoint \cite{ni-etal-2022-sentence,chen2022generate}. We use $r_1$ and $r_2$ as our labeled data with relational weights 1.0 and 0.5, respectively. We train our model for 3 epochs and evaluate every 125 training steps on the development set of STS-B to keep the best-performed checkpoint.
        
        
    \begin{table}[t]
        \centering
        \begin{adjustbox}{width=0.48\textwidth,center}
        \begin{tabular}{ l  c  c  c  c  c }
        \toprule
        \textbf{$\tau$} & 0.001 & 0.01 & 0.05 & 0.1 & 1.0 \\
        \midrule
        \textbf{Avg. STS} & 80.74 & 81.73 & 82.18 & 77.00 & 54.88 \\
        \bottomrule
        \end{tabular}
        \end{adjustbox}
        \caption{Ablation study on temperature $\tau$.}
        \label{tab:8}
    \end{table}

    \begin{table*}[thb]
        \centering
        \begin{adjustbox}{width=1.0\textwidth,center}
        \begin{tabular}{ l | c | c | c  c  c | c  c  c  c  c | c }
        \toprule
        \multirow{2}{*}{\textbf{Model}} & \multirow{2}{*}{\textbf{AskU.}} & \multirow{2}{*}{\textbf{CQADup.}} & \multicolumn{3}{c |}{\textbf{TwitterP.}} & \multicolumn{5}{c |}{\textbf{SciDocs}} & \multirow{2}{*}{\textbf{Avg.}} \\
        & & & \textbf{TURL} & \textbf{PIT} & \textbf{Avg.} & \textbf{Cite} & \textbf{CC} & \textbf{CR} & \textbf{CV} & \textbf{Avg.} &  \\
        \midrule
        GloVe (avg.) & 51.0 & 10.0 & 70.1 & 52.1 & 61.1 & 58.8 & 60.6 & 64.2 & 65.4 & 62.2 & 46.1 \\
        Sent2Vec & 49.0 & 3.2 & 47.5 & 39.9 & 43.7 & 61.6 & 66.0 & 66.1 & 66.7 & 65.1 & 40.2 \\
        BM25 & 53.4 & 13.3 & 71.9 & 70.5 & 71.2 & 58.9 & 61.3 & 67.3 & 66.9 & 63.6 & 50.4 \\
        BERT\textsubscript{\texttt{base}} & 48.5 & 6.5 & 69.1 & 61.7 & 65.4 & 59.4 & 65.1 & 65.4 & 68.6 & 64.6 & 46.3 \\
        \textsuperscript{*}SimCSE - BERT\textsubscript{\texttt{base}} & \textbf{55.9} & 12.4 & 74.5 & 62.5 & 68.5 & 62.5 & 65.1 & 67.7 & 67.6 & 65.7 & 50.6 \\
        \midrule
        SBERT\textsubscript{\texttt{base}} & 53.4 & 11.8 & 75.4 & 69.9 & 72.7 & 66.8 & 70.0 & 70.7 & 72.8 & 70.1 & 52.0 \\
        SimCSE - BERT\textsubscript{\texttt{base}} & 53.5 & 12.2 & 76.1 & 69.7 & 72.9 & 66.8 & 69.5 & \textbf{72.1} & 70.7 & 69.8 & 52.1  \\
        \midrule
        RSE - BERT\textsubscript{\texttt{base}} & 54.8 & \textbf{13.7} & \textbf{77.4} & \textbf{73.0} & \textbf{75.2} & \textbf{67.6} & \textbf{71.0} & 71.8 & \textbf{73.6} & \textbf{71.0} & \textbf{53.7} \\
        RSE - BERT\textsubscript{\texttt{large}} & 56.2 & 13.9 & 76.7 & 76.8 & 76.7 & 68.6 & 71.9 & 72.5 & 74.5 & 71.9 & 54.7 \\
        \midrule
        RSE - RoBERTa\textsubscript{\texttt{base}} & 56.2 & 13.3 & 74.8 & 74.8 & 74.8 & 66.9 & 69.4 & 70.2 & 72.5 & 69.7 & 53.5 \\
        RSE - RoBERTa\textsubscript{\texttt{large}} & \textbf{58.0} & \textbf{15.2} & 76.2 & \textbf{78.8} & \textbf{77.5} & \textbf{69.3} & \textbf{72.4} & \textbf{72.9} & \textbf{74.8} & \textbf{72.4} & \textbf{55.8} \\
        \bottomrule
        \end{tabular}
        \end{adjustbox}
        \caption{Results on domain-specific tasks. The out-of-the-box results of unsupervised and supervised approaches are shown. The first and second blocks show unsupervised and supervised methods, respectively. \textsuperscript{*} refers to unsupervised SimCSE with dropout augmentation. Our proposed RSE outperforms all benchmarking approaches.}
        \label{tab:5}
    \end{table*}

    \begin{figure}[t]
        \centering
         \includegraphics[width=0.49\textwidth]{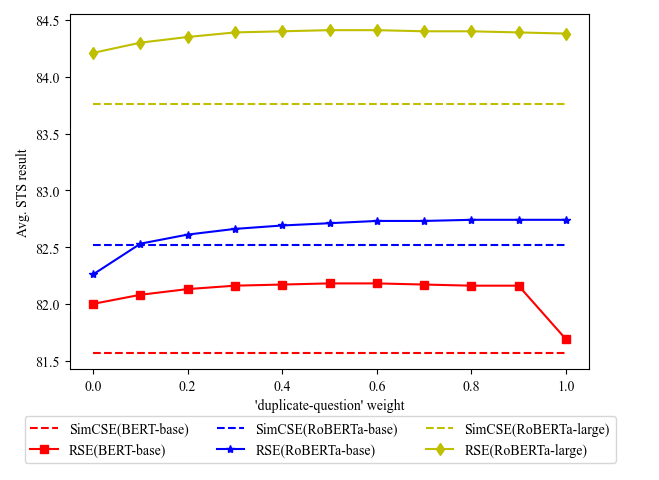}
        \caption{Performance comparison using different weights for `duplicate-question' relation for STS tasks.}
        \label{fig:RSE-weight}
    \end{figure}

    \subsection{Main Results}
    
        Table~\ref{tab:1} shows the evaluation results on 7 standard STS datasets. RSE can outperform all previous models in the supervised setting and improve the averaged STS result from 81.57\% to 82.18\% for Spearman's correlation with BERT\textsubscript{\texttt{base}} model. A further performance boost is witnessed with better pre-training checkpoints and larger model sizes. The averaged STS result with RoBERTa\textsubscript{\texttt{large}} model reaches 84.41\%. We then experiment with five random seeds to test training stability and report the standard derivation. Results show that our method is robust against randomness. The performance on the STR-2022 dataset is shown in Table~\ref{tab:3}. We observe the same result patterns, and RSE can outperform the supervised SimCSE model.

        \noindent\textbf{Effect of Temperature}. We study the influence of different $\tau$ values, and the result is shown in Table~\ref{tab:8}. $\tau=0.05$ provides the best result, and we use the same setting for all experiments.

        \noindent\textbf{Effect of Relational Weight}. We examine the performance by adjusting the weight of the relation `duplication-question' and the results are shown in Figure~\ref{fig:RSE-weight}. We witness that the avg. STS improved using the similarity score of both `duplicate question' and `entailment' relations. This phenomenon generalizes to different pre-trained models. For BERT\textsubscript{\texttt{base}} and RoBERTa\textsubscript{\texttt{large}} model, RSE can outperform others by only using the `entailment' relation. The model can perform better when involving more relational scores, which we leave for future exploration.


    

    

\section{Experiments on USEB Tasks}
\label{sec:useb}

    As there are some limitations in domain coverage of STS tasks \cite{wang-etal-2021-tsdae-using,wang2022just}, we also experiment with Universal Sentence Embedding Benchmark (USEB) proposed in \citet{wang-etal-2021-tsdae-using}. It contains four tasks from heterogeneous domains, including Re-Ranking (RR), Information Retrieval (IR) and Paraphrase Identification (PI). The datasets are AskUbuntu (RR) \cite{lei-etal-2016-semi}, CQADupStack (IR) \cite{cqadupstack}, TwitterPara (PI) \cite{xu-etal-2015-semeval,lan-etal-2017-continuously}, and SciDocs (RR) \cite{cohan-etal-2020-specter}.

    \begin{figure}[t]
        \centering
         \includegraphics[width=0.5\textwidth]{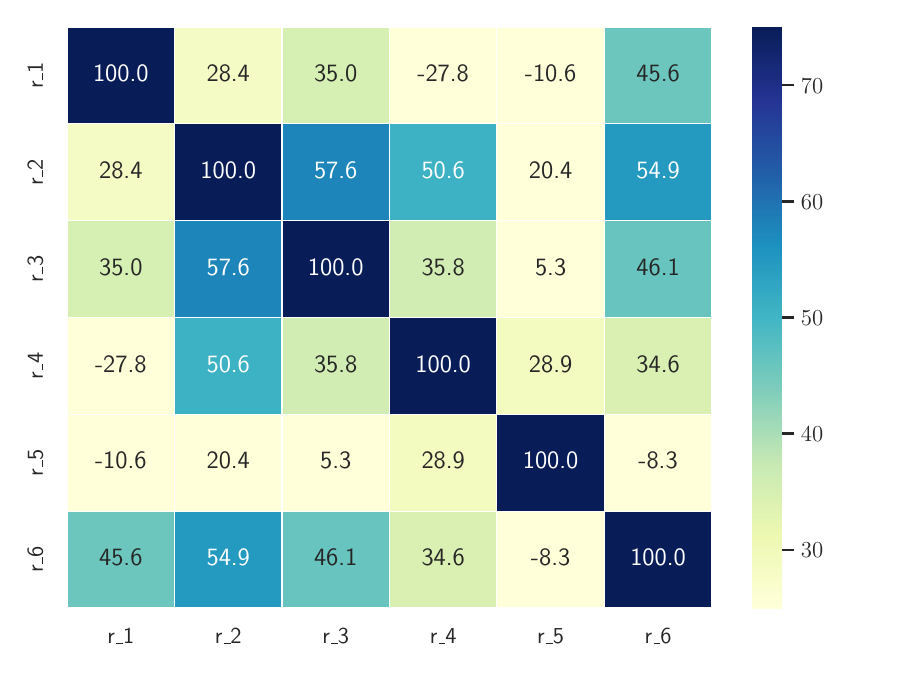}
        \caption{The similarity between relational embeddings. The relations $r_{1 \sim 6}$ are \emph{entailment, duplicate-question, paraphrase, same-caption, qa-entailment, same-sent.}}
        \label{fig:REL-SIM}
    \end{figure}

    \subsection{Settings}

        Similarly, we experiment with four different pre-trained checkpoints BERT\textsubscript{\texttt{base,large}} and RoBERTa\textsubscript{\texttt{base,large}}. We compare with different benchmarking including unsupervised methods: GloVe, Sent2Vec \cite{pagliardini-etal-2018-unsupervised}, BM25 \cite{robertson1995okapi}, BERT, unsupervised SimCSE, and supervised methods: SBERT \cite{reimers-gurevych-2019-sentence}, and SimCSE \cite{gao2021simcse}. We did not compare with other in-domain adaption-based methods for fair comparison because RSE is an off-the-shelf method. For training, we include all relational data ($r_{1\sim6}$ in Sec.~\ref{subsec:supervised-signal}) and use the similarity score from one specific relation for each task. We train our model for 3 epochs and evaluate every 1000 training steps on the development set of USEB to keep the best-performed checkpoint.

    \begin{table*}[thb]
        \centering
        \begin{adjustbox}{width=0.97\textwidth,center}
        \begin{tabular}{ l  c  c  c  c  c  c  c | c }
        \toprule
        \textbf{Model} & \textbf{MR} & \textbf{CR} & \textbf{SUBJ} & \textbf{MPQA} & \textbf{SST} & \textbf{TREC} & \textbf{MRPC} & \textbf{Avg.} \\
        \midrule \midrule
        \multicolumn{9}{c}{\textit{Unsupervised methods}} \\
        \midrule
        GloVe (avg.) & 55.14 & 70.66 & 59.73 & 68.25 & 63.66 & 58.02 & 53.76 & 61.32 \\
        BERT\textsubscript{\texttt{base}} (avg.) & 78.66 & 86.25 & 94.37 & 88.66 & 84.40 & 92.80 & 69.54 & 84.94 \\
        SimCSE - BERT\textsubscript{\texttt{base}} & 81.18 & 86.46 & 94.45 & 88.88 & 85.50 & 89.80 & 74.43 & 85.81 \\
        SimCSE - RoBERTa\textsubscript{\texttt{base}} & 81.04 & 87.74 & 93.28 & 86.94 & 86.60 & 84.60 & 73.68 & 84.84 \\
        \midrule
        \multicolumn{9}{c}{\textit{Supervised methods}} \\
        \midrule
        InferSent - GloVe & 81.57 & 86.54 & 92.50 & 90.38 & 84.18 & 88.20 & 75.77 & 85.59 \\
        USE & 80.09 & 85.19 & 93.98 & 86.70 & 86.38 & \textbf{93.20} & 70.14 & 85.10 \\
        SBERT\textsubscript{\texttt{base}} & \textbf{84.64} & \textbf{89.43} & 94.39 & 89.86 & \textbf{88.96} & 89.60 & 76.00 & 87.41 \\
        SimCSE - BERT\textsubscript{\texttt{base}} & 82.69 & 89.25 & 94.81 & 89.59 & 87.31 & 88.40 & 73.51 & 86.51 
        \\
        \quad \quad w/ MLM & 82.68 & 88.88 & 94.52 & 89.82 & 88.41 & 87.60 & 76.12 & 86.86 \\
        \midrule
        RSE - BERT\textsubscript{\texttt{base}} & 82.35 & 89.01 & \textbf{95.33} & \textbf{90.56} & 88.36 & 93.00 & \textbf{77.39} & \textbf{88.00} \\
        RSE - BERT\textsubscript{\texttt{large}} & 84.25 & 90.49 & 95.70 & 90.51 & 90.17 & 95.20 & 76.70 & 89.04 \\
        \midrule
        RSE - RoBERTa\textsubscript{\texttt{base}} & 84.52 & \textbf{91.21} & 94.39 & 90.21 & \textbf{91.49} & 91.40 & 77.10 & 88.62 \\
        RSE - RoBERTa\textsubscript{\texttt{large}} & \textbf{86.29} & \textbf{91.21} & \textbf{95.17} & \textbf{91.16} & 91.38 & \textbf{95.40} & \textbf{78.20} & \textbf{89.83} \\
        \bottomrule
        \end{tabular}
        \end{adjustbox}
        \caption{Results on 7 transfer tasks. Both unsupervised and supervised methods are compared. Two versions of SimCSE are compared. The one with an additional mask language model (MLM) objective shows better performance. RSE outperforms other approaches on average.}
        \label{tab:4}
    \end{table*}
        
    \subsection{Main Results}

        Table~\ref{tab:5} shows the evaluation result on 4 USEB tasks. Among the benchmarking methods, we found that unsupervised approaches outperform supervised approaches in AskUbuntu and CQADupStack tasks. The unsupervised SimCSE and BM25 achieve the best result on AskUbuntu and CQADupStack, respectively. It shows the heterogeneous character of USEB task and the previous supervised methods do not generalize well to various domains.

        In contrast, RSE performs the best on 3 of the 4 tasks and achieves the best overall performance. The average result improves from 52.1 to 53.7 with BERT\textsubscript{\texttt{base}} model. For AskUbuntu and CQADupStack tasks, we find that the unsupervised SimCSE method achieves the best performance, even outperforming the supervised ones. RSE framework aligns with this finding that the best re-ranking and retrieval results are obtained by the similarity inferred from the `same-sent' ($r_6$) relation. The TwitterPara dataset is for identifying whether two sentences are paraphrased or not. In RSE, we find the `paraphrase' relation ($r_3$) trained on synthetic ParaNMT dataset works the best, which aligns with our intuition. Similarly, better performance is achieved with a larger pre-trained model as weight initialization.
        The USEB experiments verify the RSE's capability in modelling multi-source relational sentence-pair data. The relational-aware similarity scores can be flexibly applied to heterogeneous domain tasks.


    \noindent\textbf{Relation Similarity}. We obtained 6 relation embeddings in our experiments after training with multi-source relational data. To study the similarity between relations, we compute the cosine similarity between pairwise relational embeddings, and the result is shown in Figure~\ref{fig:REL-SIM}. From the results, we can witness the more similar relation pairs are \emph{entailment} \& \emph{same-sent}, \emph{duplicate-question} \& \emph{paraphrase}, \emph{duplicate-question} \& \emph{same-caption}, \emph{duplicate-question} \& \emph{same-sent}, and \emph{paraphrase} \& \emph{same-sent}. The similar relations are more or less corresponding to sentence similarity, and we can see that the learned relational embeddings capture the relational information well. The \emph{qa-entailment} relation models the relationship between question and answer pairs. Therefore, we can see it is indeed less similar to all other relations.

\section{Experiments on Transfer Tasks}

    Besides the semantic similarity, ranking, and retrieval tasks, we also experiment with the following transfer tasks: MR~\cite{pang-lee-2005-seeing}, CR~\cite{hu2004mining}, SUBJ~\cite{pang-lee-2004-sentimental}, MPQA~\cite{wiebe2005annotating}, SST~\cite{socher-etal-2013-recursive}, TREC~\cite{voorhees2000building} and MRPC~\cite{dolan-brockett-2005-automatically}. Because the RSE model captures richer information by learning from multi-source relational data, we expect it generalizes better to different classification tasks.

    \subsection{Settings}

        Similarly, we experiment with four different initialization checkpoints. The averaged \texttt{[CLS]} token representation from the last five layers is taken as the sentence representation ($\mathbf{h_i}$)~\cite{wang2020sbert}. The sentence encoder is used as a feature extractor. An MLP-based classifier is trained with sentence embeddings as the input. The default setting from the SentEval package is deployed \cite{conneau-kiela-2018-senteval}.

        We leverage the relations of \emph{entailment} and \emph{paraphrase} for training. As sentence embedding is used as the input to classifiers, relation embeddings are not incorporated in transfer task experiments. All models are trained for 3 epochs. We evaluate the model on the development set every 500 training steps and keep the best-performed checkpoint for final evaluation. We compare RSE with different competitive methods including InferSent~\cite{conneau-etal-2017-supervised}, USE~\cite{cer-etal-2018-universal}, SBERT~\cite{reimers-gurevych-2019-sentence}, and SimCSE~\cite{gao2021simcse}.

        

    \subsection{Results and Analysis}    
        
        Table~\ref{tab:4} shows the evaluation results on 7 transfer tasks. The RSE model with BERT\textsubscript{\texttt{base}} architecture achieves an average score of 88.0, outperforming all benchmarked methods. SBERT outperforms SimCSE in 6 of 7 tasks. We infer that the classification-based training objective makes SBERT more transferable to transfer tasks. Even though the contrastive learning approach outperforms in semantic similarity tasks, it does not generalize well with additional trainable classifiers. In contrast, RSE is a contrastive-based approach, which already showed superior in semantic similarity tasks. Due to its learning capability from multi-source relational data, the learned sentence embeddings have better generalizability than single-source supervised learning. It leads to better transferability of the sentence embedding obtained from RSE models. Similar to previous findings, the performance of RSE improves with larger and better pre-trained checkpoints as initialization.
        Even though RSE is mainly designed for flexible semantic matching with multiple relational similarity scores, the learn sentence embedding captures richer semantic meanings for various transfer tasks.

\section{Conclusion and Future Work}

    In this work, we propose RSE, a new sentence embedding paradigm with explicit relation modelling by translation operation. RSE can learn from multi-source relational data and provide interpretable relational similarity scores. Experiments on semantic similarity, domain-specific, and transfer tasks show that RSE can outperform SOTA approaches, verifying its flexibility and effectiveness. Relational sentence embedding can broaden the concept of conventional similarity-based sentence embedding and can be extended for different applications.

    For future work, first, with more sentence relational data available, the relational sentence embedding can become more competitive and generalizable to different domains. The collection of sentence-level relational data has great potential. Second, more complex relation modelling methods can be incorporated into handling heterogeneous relations. Besides the translation operation in this work, other relation modelling functions, including rotation and semantic matching, can be further explored. Third, this work mainly focuses on supervised relation signals, and unsupervised relational sentence embedding still needs to be explored.

    
\section*{Limitations}

    The first limitation of RSE comes from data sparsity. Unlike word-level relational data, the variety of sentences is much larger than words. Therefore, even though we can collect more and more relational data on sentence pairs, it is generally hard to become densely connected between sentences. To alleviate this issue and generate sustainable sentence relational data, we should design automatic tools for sentence relation labelling with human-in-the-loop supervision.

    Second, unsupervised sentence embedding also shows decent performance on semantic textual similarity tasks, and the focus is mainly on the design of unsupervised contrastive samples. Therefore, exploring different unsupervised view-augmentation techniques in relational sentence embedding remains an open question. In this work, we only study incorporate dropout augmentation without much discussion on other techniques like pseudo-labelling \cite{chen2022generate}, word drop, word swap, and word repetition \cite{yan2021consert}.

    For relational data modelling, complex operations \cite{sun2019rotate,trouillon2016complex} for relation modelling are introduced and proven effective in knowledge graph completion. However, it is not straightforward to incorporate complex operations with pre-trained semantic representations. The study of RSE in complex space modelling remains unexplored.

\section*{Acknowledgement}

    This work is supported by the Agency for Science, Technology and Research (A*STAR) under its AME Programmatic Funding Scheme (Project No. A18A2b0046), Singapore, and by National Natural Science Foundation of China (Grant No. 62271432).




    

    



    
    

\bibliography{anthology,custom}

\begin{thebibliography}{60}
\expandafter\ifx\csname natexlab\endcsname\relax\def\natexlab#1{#1}\fi

\bibitem[{Abdalla et~al.(2021)Abdalla, Vishnubhotla, and
  Mohammad}]{abdalla2021makes}
Mohamed Abdalla, Krishnapriya Vishnubhotla, and Saif~M Mohammad. 2021.
\newblock \href {https://arxiv.org/abs/2110.04845} {What makes sentences
  semantically related: A textual relatedness dataset and empirical study}.
\newblock \emph{arXiv preprint arXiv:2110.04845}.

\bibitem[{Agirre et~al.(2015)Agirre, Banea, Cardie, Cer, Diab, Gonzalez-Agirre,
  Guo, Lopez-Gazpio, Maritxalar, Mihalcea, Rigau, Uria, and
  Wiebe}]{agirre-etal-2015-semeval}
Eneko Agirre, Carmen Banea, Claire Cardie, Daniel Cer, Mona Diab, Aitor
  Gonzalez-Agirre, Weiwei Guo, I{\~n}igo Lopez-Gazpio, Montse Maritxalar, Rada
  Mihalcea, German Rigau, Larraitz Uria, and Janyce Wiebe. 2015.
\newblock \href {https://doi.org/10.18653/v1/S15-2045} {{S}em{E}val-2015 task
  2: Semantic textual similarity, {E}nglish, {S}panish and pilot on
  interpretability}.
\newblock In \emph{Proceedings of the 9th International Workshop on Semantic
  Evaluation ({S}em{E}val 2015)}, pages 252--263, Denver, Colorado. Association
  for Computational Linguistics.

\bibitem[{Agirre et~al.(2014)Agirre, Banea, Cardie, Cer, Diab, Gonzalez-Agirre,
  Guo, Mihalcea, Rigau, and Wiebe}]{agirre-etal-2014-semeval}
Eneko Agirre, Carmen Banea, Claire Cardie, Daniel Cer, Mona Diab, Aitor
  Gonzalez-Agirre, Weiwei Guo, Rada Mihalcea, German Rigau, and Janyce Wiebe.
  2014.
\newblock \href {https://doi.org/10.3115/v1/S14-2010} {{S}em{E}val-2014 task
  10: Multilingual semantic textual similarity}.
\newblock In \emph{Proceedings of the 8th International Workshop on Semantic
  Evaluation ({S}em{E}val 2014)}, pages 81--91, Dublin, Ireland. Association
  for Computational Linguistics.

\bibitem[{Agirre et~al.(2016)Agirre, Banea, Cer, Diab, Gonzalez-Agirre,
  Mihalcea, Rigau, and Wiebe}]{agirre-etal-2016-semeval}
Eneko Agirre, Carmen Banea, Daniel Cer, Mona Diab, Aitor Gonzalez-Agirre, Rada
  Mihalcea, German Rigau, and Janyce Wiebe. 2016.
\newblock \href {https://doi.org/10.18653/v1/S16-1081} {{S}em{E}val-2016 task
  1: Semantic textual similarity, monolingual and cross-lingual evaluation}.
\newblock In \emph{Proceedings of the 10th International Workshop on Semantic
  Evaluation ({S}em{E}val-2016)}, pages 497--511, San Diego, California.
  Association for Computational Linguistics.

\bibitem[{Agirre et~al.(2012)Agirre, Cer, Diab, and
  Gonzalez-Agirre}]{agirre-etal-2012-semeval}
Eneko Agirre, Daniel Cer, Mona Diab, and Aitor Gonzalez-Agirre. 2012.
\newblock \href {https://aclanthology.org/S12-1051} {{S}em{E}val-2012 task 6: A
  pilot on semantic textual similarity}.
\newblock In \emph{*{SEM} 2012: The First Joint Conference on Lexical and
  Computational Semantics {--} Volume 1: Proceedings of the main conference and
  the shared task, and Volume 2: Proceedings of the Sixth International
  Workshop on Semantic Evaluation ({S}em{E}val 2012)}, pages 385--393,
  Montr{\'e}al, Canada. Association for Computational Linguistics.

\bibitem[{Agirre et~al.(2013)Agirre, Cer, Diab, Gonzalez-Agirre, and
  Guo}]{agirre-etal-2013-sem}
Eneko Agirre, Daniel Cer, Mona Diab, Aitor Gonzalez-Agirre, and Weiwei Guo.
  2013.
\newblock \href {https://aclanthology.org/S13-1004} {*{SEM} 2013 shared task:
  Semantic textual similarity}.
\newblock In \emph{Second Joint Conference on Lexical and Computational
  Semantics (*{SEM}), Volume 1: Proceedings of the Main Conference and the
  Shared Task: Semantic Textual Similarity}, pages 32--43, Atlanta, Georgia,
  USA. Association for Computational Linguistics.

\bibitem[{Balazevic et~al.(2019)Balazevic, Allen, and
  Hospedales}]{balazevic-etal-2019-tucker}
Ivana Balazevic, Carl Allen, and Timothy Hospedales. 2019.
\newblock \href {https://doi.org/10.18653/v1/D19-1522} {{T}uck{ER}: Tensor
  factorization for knowledge graph completion}.
\newblock In \emph{Proceedings of the 2019 Conference on Empirical Methods in
  Natural Language Processing and the 9th International Joint Conference on
  Natural Language Processing (EMNLP-IJCNLP)}, pages 5185--5194, Hong Kong,
  China. Association for Computational Linguistics.

\bibitem[{Bordes et~al.(2013)Bordes, Usunier, Garcia-Duran, Weston, and
  Yakhnenko}]{bordes2013translating}
Antoine Bordes, Nicolas Usunier, Alberto Garcia-Duran, Jason Weston, and Oksana
  Yakhnenko. 2013.
\newblock Translating embeddings for modeling multi-relational data.
\newblock \emph{Advances in neural information processing systems}, 26.

\bibitem[{Bowman et~al.(2015)Bowman, Angeli, Potts, and
  Manning}]{bowman-etal-2015-large}
Samuel~R. Bowman, Gabor Angeli, Christopher Potts, and Christopher~D. Manning.
  2015.
\newblock \href {https://doi.org/10.18653/v1/D15-1075} {A large annotated
  corpus for learning natural language inference}.
\newblock In \emph{Proceedings of the 2015 Conference on Empirical Methods in
  Natural Language Processing}, pages 632--642, Lisbon, Portugal. Association
  for Computational Linguistics.

\bibitem[{Camacho-Collados et~al.(2019)Camacho-Collados, Espinosa~Anke, and
  Schockaert}]{camacho-collados-etal-2019-relational}
Jose Camacho-Collados, Luis Espinosa~Anke, and Steven Schockaert. 2019.
\newblock \href {https://doi.org/10.18653/v1/P19-1318} {Relational word
  embeddings}.
\newblock In \emph{Proceedings of the 57th Annual Meeting of the Association
  for Computational Linguistics}, pages 3286--3296, Florence, Italy.
  Association for Computational Linguistics.

\bibitem[{Cer et~al.(2017)Cer, Diab, Agirre, Lopez-Gazpio, and
  Specia}]{cer-etal-2017-semeval}
Daniel Cer, Mona Diab, Eneko Agirre, I{\~n}igo Lopez-Gazpio, and Lucia Specia.
  2017.
\newblock \href {https://doi.org/10.18653/v1/S17-2001} {{S}em{E}val-2017 task
  1: Semantic textual similarity multilingual and crosslingual focused
  evaluation}.
\newblock In \emph{Proceedings of the 11th International Workshop on Semantic
  Evaluation ({S}em{E}val-2017)}, pages 1--14, Vancouver, Canada. Association
  for Computational Linguistics.

\bibitem[{Cer et~al.(2018)Cer, Yang, Kong, Hua, Limtiaco, St.~John, Constant,
  Guajardo-Cespedes, Yuan, Tar, Strope, and Kurzweil}]{cer-etal-2018-universal}
Daniel Cer, Yinfei Yang, Sheng-yi Kong, Nan Hua, Nicole Limtiaco, Rhomni
  St.~John, Noah Constant, Mario Guajardo-Cespedes, Steve Yuan, Chris Tar,
  Brian Strope, and Ray Kurzweil. 2018.
\newblock \href {https://doi.org/10.18653/v1/D18-2029} {Universal sentence
  encoder for {E}nglish}.
\newblock In \emph{Proceedings of the 2018 Conference on Empirical Methods in
  Natural Language Processing: System Demonstrations}, pages 169--174,
  Brussels, Belgium. Association for Computational Linguistics.

\bibitem[{Chen et~al.(2020)Chen, Kornblith, Norouzi, and
  Hinton}]{chen2020simple}
Ting Chen, Simon Kornblith, Mohammad Norouzi, and Geoffrey Hinton. 2020.
\newblock A simple framework for contrastive learning of visual
  representations.
\newblock In \emph{International conference on machine learning}, pages
  1597--1607. PMLR.

\bibitem[{Chen et~al.(2022)Chen, Zhang, Wang, Liu, and Li}]{chen2022generate}
Yiming Chen, Yan Zhang, Bin Wang, Zuozhu Liu, and Haizhou Li. 2022.
\newblock \href {https://aclanthology.org/2022.emnlp-main.558} {Generate,
  discriminate and contrast: A semi-supervised sentence representation learning
  framework}.
\newblock In \emph{Proceedings of the 2022 Conference on Empirical Methods in
  Natural Language Processing}, pages 8150--8161, Abu Dhabi, United Arab
  Emirates. Association for Computational Linguistics.

\bibitem[{Cohan et~al.(2020)Cohan, Feldman, Beltagy, Downey, and
  Weld}]{cohan-etal-2020-specter}
Arman Cohan, Sergey Feldman, Iz~Beltagy, Doug Downey, and Daniel Weld. 2020.
\newblock \href {https://doi.org/10.18653/v1/2020.acl-main.207} {{SPECTER}:
  Document-level representation learning using citation-informed transformers}.
\newblock In \emph{Proceedings of the 58th Annual Meeting of the Association
  for Computational Linguistics}, pages 2270--2282, Online. Association for
  Computational Linguistics.

\bibitem[{Conneau and Kiela(2018)}]{conneau-kiela-2018-senteval}
Alexis Conneau and Douwe Kiela. 2018.
\newblock \href {https://aclanthology.org/L18-1269} {{S}ent{E}val: An
  evaluation toolkit for universal sentence representations}.
\newblock In \emph{Proceedings of the Eleventh International Conference on
  Language Resources and Evaluation ({LREC} 2018)}, Miyazaki, Japan. European
  Language Resources Association (ELRA).

\bibitem[{Conneau et~al.(2017)Conneau, Kiela, Schwenk, Barrault, and
  Bordes}]{conneau-etal-2017-supervised}
Alexis Conneau, Douwe Kiela, Holger Schwenk, Lo{\"\i}c Barrault, and Antoine
  Bordes. 2017.
\newblock \href {https://doi.org/10.18653/v1/D17-1070} {Supervised learning of
  universal sentence representations from natural language inference data}.
\newblock In \emph{Proceedings of the 2017 Conference on Empirical Methods in
  Natural Language Processing}, pages 670--680, Copenhagen, Denmark.
  Association for Computational Linguistics.

\bibitem[{Dettmers et~al.(2018)Dettmers, Minervini, Stenetorp, and
  Riedel}]{dettmers2018convolutional}
Tim Dettmers, Pasquale Minervini, Pontus Stenetorp, and Sebastian Riedel. 2018.
\newblock Convolutional 2d knowledge graph embeddings.
\newblock In \emph{Proceedings of the AAAI conference on artificial
  intelligence}, volume~32.

\bibitem[{Devlin et~al.(2019)Devlin, Chang, Lee, and
  Toutanova}]{devlin-etal-2019-bert}
Jacob Devlin, Ming-Wei Chang, Kenton Lee, and Kristina Toutanova. 2019.
\newblock \href {https://doi.org/10.18653/v1/N19-1423} {{BERT}: Pre-training of
  deep bidirectional transformers for language understanding}.
\newblock In \emph{Proceedings of the 2019 Conference of the North {A}merican
  Chapter of the Association for Computational Linguistics: Human Language
  Technologies, Volume 1 (Long and Short Papers)}, pages 4171--4186,
  Minneapolis, Minnesota. Association for Computational Linguistics.

\bibitem[{Dolan and Brockett(2005)}]{dolan-brockett-2005-automatically}
William~B. Dolan and Chris Brockett. 2005.
\newblock \href {https://aclanthology.org/I05-5002} {Automatically constructing
  a corpus of sentential paraphrases}.
\newblock In \emph{Proceedings of the Third International Workshop on
  Paraphrasing ({IWP}2005)}.

\bibitem[{Gao et~al.(2021{\natexlab{a}})Gao, Zhang, Han, and
  Callan}]{gao-etal-2021-scaling}
Luyu Gao, Yunyi Zhang, Jiawei Han, and Jamie Callan. 2021{\natexlab{a}}.
\newblock \href {https://doi.org/10.18653/v1/2021.repl4nlp-1.31} {Scaling deep
  contrastive learning batch size under memory limited setup}.
\newblock In \emph{Proceedings of the 6th Workshop on Representation Learning
  for NLP (RepL4NLP-2021)}, pages 316--321, Online. Association for
  Computational Linguistics.

\bibitem[{Gao et~al.(2021{\natexlab{b}})Gao, Yao, and Chen}]{gao2021simcse}
Tianyu Gao, Xingcheng Yao, and Danqi Chen. 2021{\natexlab{b}}.
\newblock \href {https://doi.org/10.18653/v1/2021.emnlp-main.552} {{S}im{CSE}:
  Simple contrastive learning of sentence embeddings}.
\newblock In \emph{Proceedings of the 2021 Conference on Empirical Methods in
  Natural Language Processing}, pages 6894--6910, Online and Punta Cana,
  Dominican Republic. Association for Computational Linguistics.

\bibitem[{He et~al.(2020)He, Fan, Wu, Xie, and Girshick}]{he2020momentum}
Kaiming He, Haoqi Fan, Yuxin Wu, Saining Xie, and Ross Girshick. 2020.
\newblock Momentum contrast for unsupervised visual representation learning.
\newblock In \emph{Proceedings of the IEEE/CVF conference on computer vision
  and pattern recognition}, pages 9729--9738.

\bibitem[{Hoogeveen et~al.(2015)Hoogeveen, Verspoor, and Baldwin}]{cqadupstack}
Doris Hoogeveen, Karin~M. Verspoor, and Timothy Baldwin. 2015.
\newblock \href {https://doi.org/10.1145/2838931.2838934} {Cqadupstack: A
  benchmark data set for community question-answering research}.
\newblock In \emph{Proceedings of the 20th Australasian Document Computing
  Symposium}, ADCS '15, New York, NY, USA. Association for Computing Machinery.

\bibitem[{Hu and Liu(2004)}]{hu2004mining}
Minqing Hu and Bing Liu. 2004.
\newblock Mining and summarizing customer reviews.
\newblock In \emph{Proceedings of the tenth ACM SIGKDD international conference
  on Knowledge discovery and data mining}, pages 168--177.

\bibitem[{Janson et~al.(2021)Janson, Gogoulou, Ylip{\"a}{\"a}, Cuba~Gyllensten,
  and Sahlgren}]{janson2021semantic}
Sverker Janson, Evangelina Gogoulou, Erik Ylip{\"a}{\"a}, Amaru
  Cuba~Gyllensten, and Magnus Sahlgren. 2021.
\newblock Semantic re-tuning with contrastive tension.
\newblock In \emph{International Conference on Learning Representations, 2021}.

\bibitem[{Kiros et~al.(2015)Kiros, Zhu, Salakhutdinov, Zemel, Urtasun,
  Torralba, and Fidler}]{kiros2015skip}
Ryan Kiros, Yukun Zhu, Russ~R Salakhutdinov, Richard Zemel, Raquel Urtasun,
  Antonio Torralba, and Sanja Fidler. 2015.
\newblock \href {https://arxiv.org/abs/1506.06726} {Skip-thought vectors}.
\newblock \emph{Advances in neural information processing systems}, 28.

\bibitem[{Lan et~al.(2017)Lan, Qiu, He, and Xu}]{lan-etal-2017-continuously}
Wuwei Lan, Siyu Qiu, Hua He, and Wei Xu. 2017.
\newblock \href {https://doi.org/10.18653/v1/D17-1126} {A continuously growing
  dataset of sentential paraphrases}.
\newblock In \emph{Proceedings of the 2017 Conference on Empirical Methods in
  Natural Language Processing}, pages 1224--1234, Copenhagen, Denmark.
  Association for Computational Linguistics.

\bibitem[{Lei et~al.(2016)Lei, Joshi, Barzilay, Jaakkola, Tymoshenko,
  Moschitti, and M{\`a}rquez}]{lei-etal-2016-semi}
Tao Lei, Hrishikesh Joshi, Regina Barzilay, Tommi Jaakkola, Kateryna
  Tymoshenko, Alessandro Moschitti, and Llu{\'\i}s M{\`a}rquez. 2016.
\newblock \href {https://doi.org/10.18653/v1/N16-1153} {Semi-supervised
  question retrieval with gated convolutions}.
\newblock In \emph{Proceedings of the 2016 Conference of the North {A}merican
  Chapter of the Association for Computational Linguistics: Human Language
  Technologies}, pages 1279--1289, San Diego, California. Association for
  Computational Linguistics.

\bibitem[{Lin et~al.(2015)Lin, Liu, Sun, Liu, and Zhu}]{lin2015learning}
Yankai Lin, Zhiyuan Liu, Maosong Sun, Yang Liu, and Xuan Zhu. 2015.
\newblock \href {https://ojs.aaai.org/index.php/AAAI/article/view/9491}
  {Learning entity and relation embeddings for knowledge graph completion}.
\newblock In \emph{Twenty-ninth AAAI conference on artificial intelligence}.

\bibitem[{Liu et~al.(2019)Liu, Ott, Goyal, Du, Joshi, Chen, Levy, Lewis,
  Zettlemoyer, and Stoyanov}]{liu2019roberta}
Yinhan Liu, Myle Ott, Naman Goyal, Jingfei Du, Mandar Joshi, Danqi Chen, Omer
  Levy, Mike Lewis, Luke Zettlemoyer, and Veselin Stoyanov. 2019.
\newblock \href {https://arxiv.org/abs/1907.11692} {Roberta: A robustly
  optimized bert pretraining approach}.
\newblock \emph{arXiv preprint arXiv:1907.11692}.

\bibitem[{Marelli et~al.(2014)Marelli, Menini, Baroni, Bentivogli, Bernardi,
  and Zamparelli}]{marelli-etal-2014-sick}
Marco Marelli, Stefano Menini, Marco Baroni, Luisa Bentivogli, Raffaella
  Bernardi, and Roberto Zamparelli. 2014.
\newblock \href
  {http://www.lrec-conf.org/proceedings/lrec2014/pdf/363_Paper.pdf} {A {SICK}
  cure for the evaluation of compositional distributional semantic models}.
\newblock In \emph{Proceedings of the Ninth International Conference on
  Language Resources and Evaluation ({LREC}'14)}, pages 216--223, Reykjavik,
  Iceland. European Language Resources Association (ELRA).

\bibitem[{Miller(1995)}]{miller1995wordnet}
George~A Miller. 1995.
\newblock \href {https://aclanthology.org/H94-1111.pdf} {Wordnet: a lexical
  database for english}.
\newblock \emph{Communications of the ACM}, 38(11):39--41.

\bibitem[{Ni et~al.(2022)Ni, Hernandez~Abrego, Constant, Ma, Hall, Cer, and
  Yang}]{ni-etal-2022-sentence}
Jianmo Ni, Gustavo Hernandez~Abrego, Noah Constant, Ji~Ma, Keith Hall, Daniel
  Cer, and Yinfei Yang. 2022.
\newblock \href {https://doi.org/10.18653/v1/2022.findings-acl.146}
  {Sentence-t5: Scalable sentence encoders from pre-trained text-to-text
  models}.
\newblock In \emph{Findings of the Association for Computational Linguistics:
  ACL 2022}, pages 1864--1874, Dublin, Ireland. Association for Computational
  Linguistics.

\bibitem[{Pagliardini et~al.(2018)Pagliardini, Gupta, and
  Jaggi}]{pagliardini-etal-2018-unsupervised}
Matteo Pagliardini, Prakhar Gupta, and Martin Jaggi. 2018.
\newblock \href {https://doi.org/10.18653/v1/N18-1049} {Unsupervised learning
  of sentence embeddings using compositional n-gram features}.
\newblock In \emph{Proceedings of the 2018 Conference of the North {A}merican
  Chapter of the Association for Computational Linguistics: Human Language
  Technologies, Volume 1 (Long Papers)}, pages 528--540, New Orleans,
  Louisiana. Association for Computational Linguistics.

\bibitem[{Pang and Lee(2004)}]{pang-lee-2004-sentimental}
Bo~Pang and Lillian Lee. 2004.
\newblock \href {https://doi.org/10.3115/1218955.1218990} {A sentimental
  education: Sentiment analysis using subjectivity summarization based on
  minimum cuts}.
\newblock In \emph{Proceedings of the 42nd Annual Meeting of the Association
  for Computational Linguistics ({ACL}-04)}, pages 271--278, Barcelona, Spain.

\bibitem[{Pang and Lee(2005)}]{pang-lee-2005-seeing}
Bo~Pang and Lillian Lee. 2005.
\newblock \href {https://doi.org/10.3115/1219840.1219855} {Seeing stars:
  Exploiting class relationships for sentiment categorization with respect to
  rating scales}.
\newblock In \emph{Proceedings of the 43rd Annual Meeting of the Association
  for Computational Linguistics ({ACL}{'}05)}, pages 115--124, Ann Arbor,
  Michigan. Association for Computational Linguistics.

\bibitem[{Pennington et~al.(2014)Pennington, Socher, and
  Manning}]{pennington-etal-2014-glove}
Jeffrey Pennington, Richard Socher, and Christopher Manning. 2014.
\newblock \href {https://doi.org/10.3115/v1/D14-1162} {{G}lo{V}e: Global
  vectors for word representation}.
\newblock In \emph{Proceedings of the 2014 Conference on Empirical Methods in
  Natural Language Processing ({EMNLP})}, pages 1532--1543, Doha, Qatar.
  Association for Computational Linguistics.

\bibitem[{Reimers and Gurevych(2019)}]{reimers-gurevych-2019-sentence}
Nils Reimers and Iryna Gurevych. 2019.
\newblock \href {https://doi.org/10.18653/v1/D19-1410} {Sentence-{BERT}:
  Sentence embeddings using {S}iamese {BERT}-networks}.
\newblock In \emph{Proceedings of the 2019 Conference on Empirical Methods in
  Natural Language Processing and the 9th International Joint Conference on
  Natural Language Processing (EMNLP-IJCNLP)}, pages 3982--3992, Hong Kong,
  China. Association for Computational Linguistics.

\bibitem[{Robertson et~al.(1995)Robertson, Walker, Jones, Hancock-Beaulieu,
  Gatford et~al.}]{robertson1995okapi}
Stephen~E Robertson, Steve Walker, Susan Jones, Micheline~M Hancock-Beaulieu,
  Mike Gatford, et~al. 1995.
\newblock \href
  {https://www.microsoft.com/en-us/research/wp-content/uploads/2016/02/okapi_trec3.pdf}
  {Okapi at trec-3}.
\newblock \emph{Nist Special Publication Sp}, 109:109.

\bibitem[{Socher et~al.(2013)Socher, Perelygin, Wu, Chuang, Manning, Ng, and
  Potts}]{socher-etal-2013-recursive}
Richard Socher, Alex Perelygin, Jean Wu, Jason Chuang, Christopher~D. Manning,
  Andrew Ng, and Christopher Potts. 2013.
\newblock \href {https://aclanthology.org/D13-1170} {Recursive deep models for
  semantic compositionality over a sentiment treebank}.
\newblock In \emph{Proceedings of the 2013 Conference on Empirical Methods in
  Natural Language Processing}, pages 1631--1642, Seattle, Washington, USA.
  Association for Computational Linguistics.

\bibitem[{Su et~al.(2021)Su, Cao, Liu, and Ou}]{su2021whitening}
Jianlin Su, Jiarun Cao, Weijie Liu, and Yangyiwen Ou. 2021.
\newblock \href {https://arxiv.org/abs/2103.15316} {Whitening sentence
  representations for better semantics and faster retrieval}.
\newblock \emph{arXiv preprint arXiv:2103.15316}.

\bibitem[{Sun et~al.(2019)Sun, Deng, Nie, and Tang}]{sun2019rotate}
Zhiqing Sun, Zhi-Hong Deng, Jian-Yun Nie, and Jian Tang. 2019.
\newblock \href {https://openreview.net/forum?id=HkgEQnRqYQ} {Rotate: Knowledge
  graph embedding by relational rotation in complex space}.
\newblock In \emph{International Conference on Learning Representations}.

\bibitem[{Trouillon et~al.(2016)Trouillon, Welbl, Riedel, Gaussier, and
  Bouchard}]{trouillon2016complex}
Th{\'e}o Trouillon, Johannes Welbl, Sebastian Riedel, {\'E}ric Gaussier, and
  Guillaume Bouchard. 2016.
\newblock \href {https://dl.acm.org/doi/10.5555/3045390.3045609} {Complex
  embeddings for simple link prediction}.
\newblock In \emph{International conference on machine learning}, pages
  2071--2080. PMLR.

\bibitem[{Voorhees and Tice(2000)}]{voorhees2000building}
Ellen~M Voorhees and Dawn~M Tice. 2000.
\newblock \href {https://dl.acm.org/doi/10.1145/345508.345577} {Building a
  question answering test collection}.
\newblock In \emph{Proceedings of the 23rd annual international ACM SIGIR
  conference on Research and development in information retrieval}, pages
  200--207.

\bibitem[{Wang et~al.(2018)Wang, Singh, Michael, Hill, Levy, and
  Bowman}]{wang-etal-2018-glue}
Alex Wang, Amanpreet Singh, Julian Michael, Felix Hill, Omer Levy, and Samuel
  Bowman. 2018.
\newblock \href {https://doi.org/10.18653/v1/W18-5446} {{GLUE}: A multi-task
  benchmark and analysis platform for natural language understanding}.
\newblock In \emph{Proceedings of the 2018 {EMNLP} Workshop {B}lackbox{NLP}:
  Analyzing and Interpreting Neural Networks for {NLP}}, pages 353--355,
  Brussels, Belgium. Association for Computational Linguistics.

\bibitem[{Wang and Kuo(2020)}]{wang2020sbert}
Bin Wang and C.-C.~Jay Kuo. 2020.
\newblock {SBERT-WK}: A sentence embedding method by dissecting bert-based word
  models.
\newblock \emph{IEEE/ACM Transactions on Audio, Speech, and Language
  Processing}, 28:2146--2157.

\bibitem[{Wang et~al.(2022)Wang, Kuo, and Li}]{wang2022just}
Bin Wang, C.-C.~Jay Kuo, and Haizhou Li. 2022.
\newblock \href {https://doi.org/10.18653/v1/2022.acl-long.419} {Just rank:
  Rethinking evaluation with word and sentence similarities}.
\newblock In \emph{Proceedings of the 60th Annual Meeting of the Association
  for Computational Linguistics (Volume 1: Long Papers)}, pages 6060--6077,
  Dublin, Ireland. Association for Computational Linguistics.

\bibitem[{Wang et~al.(2021{\natexlab{a}})Wang, Wang, Huang, You, Leskovec, and
  Kuo}]{wang2021inductive}
Bin Wang, Guangtao Wang, Jing Huang, Jiaxuan You, Jure Leskovec, and C.-C.~Jay
  Kuo. 2021{\natexlab{a}}.
\newblock \href {https://ieeexplore.ieee.org/document/9534355} {Inductive
  learning on commonsense knowledge graph completion}.
\newblock In \emph{2021 International Joint Conference on Neural Networks
  (IJCNN)}, pages 1--8. IEEE.

\bibitem[{Wang et~al.(2021{\natexlab{b}})Wang, Reimers, and
  Gurevych}]{wang-etal-2021-tsdae-using}
Kexin Wang, Nils Reimers, and Iryna Gurevych. 2021{\natexlab{b}}.
\newblock \href {https://doi.org/10.18653/v1/2021.findings-emnlp.59} {{TSDAE}:
  Using transformer-based sequential denoising auto-encoderfor unsupervised
  sentence embedding learning}.
\newblock In \emph{Findings of the Association for Computational Linguistics:
  EMNLP 2021}, pages 671--688, Punta Cana, Dominican Republic. Association for
  Computational Linguistics.

\bibitem[{Wang et~al.(2016)Wang, Li, Liu, and Tang}]{wang2016text}
Zhigang Wang, Juanzi Li, Zhiyuan Liu, and Jie Tang. 2016.
\newblock \href {https://www.ijcai.org/Proceedings/16/Papers/187.pdf}
  {Text-enhanced representation learning for knowledge graph}.
\newblock In \emph{Proceedings of International joint conference on artificial
  intelligent (IJCAI)}, pages 4--17.

\bibitem[{Wiebe et~al.(2005)Wiebe, Wilson, and Cardie}]{wiebe2005annotating}
Janyce Wiebe, Theresa Wilson, and Claire Cardie. 2005.
\newblock \href {https://link.springer.com/article/10.1007/s10579-005-7880-9}
  {Annotating expressions of opinions and emotions in language}.
\newblock \emph{Language resources and evaluation}, 39(2):165--210.

\bibitem[{Wieting and Gimpel(2018)}]{wieting2017paranmt}
John Wieting and Kevin Gimpel. 2018.
\newblock \href {https://doi.org/10.18653/v1/P18-1042} {{P}ara{NMT}-50{M}:
  Pushing the limits of paraphrastic sentence embeddings with millions of
  machine translations}.
\newblock In \emph{Proceedings of the 56th Annual Meeting of the Association
  for Computational Linguistics (Volume 1: Long Papers)}, pages 451--462,
  Melbourne, Australia. Association for Computational Linguistics.

\bibitem[{Williams et~al.(2018)Williams, Nangia, and
  Bowman}]{williams-etal-2018-broad}
Adina Williams, Nikita Nangia, and Samuel Bowman. 2018.
\newblock \href {https://doi.org/10.18653/v1/N18-1101} {A broad-coverage
  challenge corpus for sentence understanding through inference}.
\newblock In \emph{Proceedings of the 2018 Conference of the North {A}merican
  Chapter of the Association for Computational Linguistics: Human Language
  Technologies, Volume 1 (Long Papers)}, pages 1112--1122, New Orleans,
  Louisiana. Association for Computational Linguistics.

\bibitem[{Wu et~al.(2022)Wu, Gao, Zang, Han, Wang, and Hu}]{wu2021esimcse}
Xing Wu, Chaochen Gao, Liangjun Zang, Jizhong Han, Zhongyuan Wang, and Songlin
  Hu. 2022.
\newblock \href {https://aclanthology.org/2022.coling-1.342} {{ES}im{CSE}:
  Enhanced sample building method for contrastive learning of unsupervised
  sentence embedding}.
\newblock In \emph{Proceedings of the 29th International Conference on
  Computational Linguistics}, pages 3898--3907, Gyeongju, Republic of Korea.
  International Committee on Computational Linguistics.

\bibitem[{Xu et~al.(2015)Xu, Callison-Burch, and Dolan}]{xu-etal-2015-semeval}
Wei Xu, Chris Callison-Burch, and Bill Dolan. 2015.
\newblock \href {https://doi.org/10.18653/v1/S15-2001} {{S}em{E}val-2015 task
  1: Paraphrase and semantic similarity in {T}witter ({PIT})}.
\newblock In \emph{Proceedings of the 9th International Workshop on Semantic
  Evaluation ({S}em{E}val 2015)}, pages 1--11, Denver, Colorado. Association
  for Computational Linguistics.

\bibitem[{Yan et~al.(2021)Yan, Li, Wang, Zhang, Wu, and Xu}]{yan2021consert}
Yuanmeng Yan, Rumei Li, Sirui Wang, Fuzheng Zhang, Wei Wu, and Weiran Xu. 2021.
\newblock \href {https://doi.org/10.18653/v1/2021.acl-long.393} {{C}on{SERT}: A
  contrastive framework for self-supervised sentence representation transfer}.
\newblock In \emph{Proceedings of the 59th Annual Meeting of the Association
  for Computational Linguistics and the 11th International Joint Conference on
  Natural Language Processing (Volume 1: Long Papers)}, pages 5065--5075,
  Online. Association for Computational Linguistics.

\bibitem[{Yang et~al.(2015)Yang, Yih, He, Gao, and Deng}]{yang2014embedding}
Bishan Yang, Wen-tau Yih, Xiaodong He, Jianfeng Gao, and Li~Deng. 2015.
\newblock \href {https://arxiv.org/abs/1412.6575} {Embedding entities and
  relations for learning and inference in knowledge bases}.
\newblock \emph{International Conference on Learning Representations}.

\bibitem[{Yao et~al.(2019)Yao, Mao, and Luo}]{yao2019kg}
Liang Yao, Chengsheng Mao, and Yuan Luo. 2019.
\newblock \href {https://arxiv.org/abs/1909.03193} {{KG-BERT}: {BERT} for
  knowledge graph completion}.
\newblock \emph{arXiv preprint arXiv:1909.03193}.

\bibitem[{Young et~al.(2014)Young, Lai, Hodosh, and
  Hockenmaier}]{young-etal-2014-image}
Peter Young, Alice Lai, Micah Hodosh, and Julia Hockenmaier. 2014.
\newblock \href {https://doi.org/10.1162/tacl_a_00166} {From image descriptions
  to visual denotations: New similarity metrics for semantic inference over
  event descriptions}.
\newblock \emph{Transactions of the Association for Computational Linguistics},
  2:67--78.

\end{thebibliography}
\bibliographystyle{acl_natbib}


\appendix

\section{Joint-Learning with Multiple Datasets}
\label{appendix:mscs}

    SimCSE~\cite{gao2021simcse} found that the positive sentence pairs extracted from \emph{entailment} relation are the most effective supervision signals in contrastive sentence representation learning. We first experiment with different supervision as positive sentence pairs. The averaged STS result is reported in Table~\ref{tab:demo1}. The result shows that using \emph{entailment} for positive and \emph{contrastive} for negatives provides the best performance. It outperforms all other supervision, including QQP, Flicker, ParaNMT, and QNLI datasets. This explains why only NLI dataset is widely used in previous work for supervised sentence representation learning \cite{conneau-etal-2017-supervised,reimers-gurevych-2019-sentence,gao2021simcse}.

    Another question comes to the stage: Can we simultaneously leverage the supervision from the above-mentioned datasets for better-supervised representation learning? An intuitive way to learn from multi-source relational data is to treat all supervised signals as positive samples in contrastive learning. Therefore, we experiment with merging all supervision data and performing contrastive sentence representation learning following \citet{gao2021simcse}. The result is shown in Table~\ref{tab:demo1}. We can tell that merging all sentence pairs for joint training will hurt the performance (77.67 -> 74.62) of STS datasets compared with a single source for positive sentence pairs.

    Based on the above experimental results, we conclude it is not a trivial task to leverage multi-source supervision signals in supervised sentence representation learning. With explicit relation modeling techniques proposed in RSE, we can effectively learn from multi-source relational data and achieve better flexibility and performance in various tasks.





    \begin{table}[htb]
        \centering
        \begin{adjustbox}{width=0.5\textwidth,center}
        \begin{tabular}{ l  c  c  c  c  c }
        \toprule
        \textbf{Relation} & \textbf{MRR} & \textbf{Hits$@$1} & \textbf{Hits$@$3} & \textbf{Hits$@$10} \\
        \midrule
        Entailment & 0.80 & 0.74 & 0.85 & 0.93 \\
        Duplicate Question & 0.83 & 0.76 & 0.88 & 0.97 \\
        Paraphrase & 0.97 & 0.95 & 0.98 & 1.00 \\
        QA Entailment & 0.85 & 0.80 & 0.90 & 0.95 \\
        Same Caption & 0.56 & 0.45 & 0.63 & 0.79 \\ \midrule
        All & 0.81 & 0.74 & 0.84 & 0.92 \\
        \bottomrule
        \end{tabular}
        \end{adjustbox}
        \caption{Performance on in-domain relation prediction. We follow the evaluation protocol of link prediction in knowledge graph completion literature.}
        \label{tab:relpred}
    \end{table}

\section{In-domain Relation Prediction}


    As the relation sentence embedding is trained in a contrastive manner, the trained model can also be used to predict the validity of new triples. Therefore, we did further experiments to test the model's performance on link prediction.

    For dataset creation, we use data from five different relations, including \emph{entailment}, \emph{duplicate-question}, \emph{paraphrase}, \emph{same-caption}, and \emph{qa-entailment}. We did not include the prediction for \emph{same-sent} relation as the prediction for the same sentence is a trivial task. We use all available data for training. For validation and testing, we randomly sampled 2,000 sentence triples each.

    We perform link prediction experiments following the literature of knowledge graph completion \cite{sun2019rotate}. The task is to predict the tail entity (sentence) given the head entity (sentence) and relation. It is a retrieval-based task, and performance is measured by the rank of the valid positive samples. Our experimental result is shown in Table~\ref{tab:relpred}. From the result, we can see that the \emph{paraphrase} relation is the easiest to predict. It is understandable because the sentences from the paraphrased data are very similar. In contrast, we see that the \emph{same-caption} relation is the hardest to be retrieved. It is because the sentence pairs are collected from the descriptions for the same image. The captions for each image can be generated based on various aspects. The relationship between large number sentence pairs is not very strong. It poses challenges on \emph{same-caption} relations in link prediction.

    \begin{figure}[htb]
        \centering
         \includegraphics[width=0.5\textwidth]{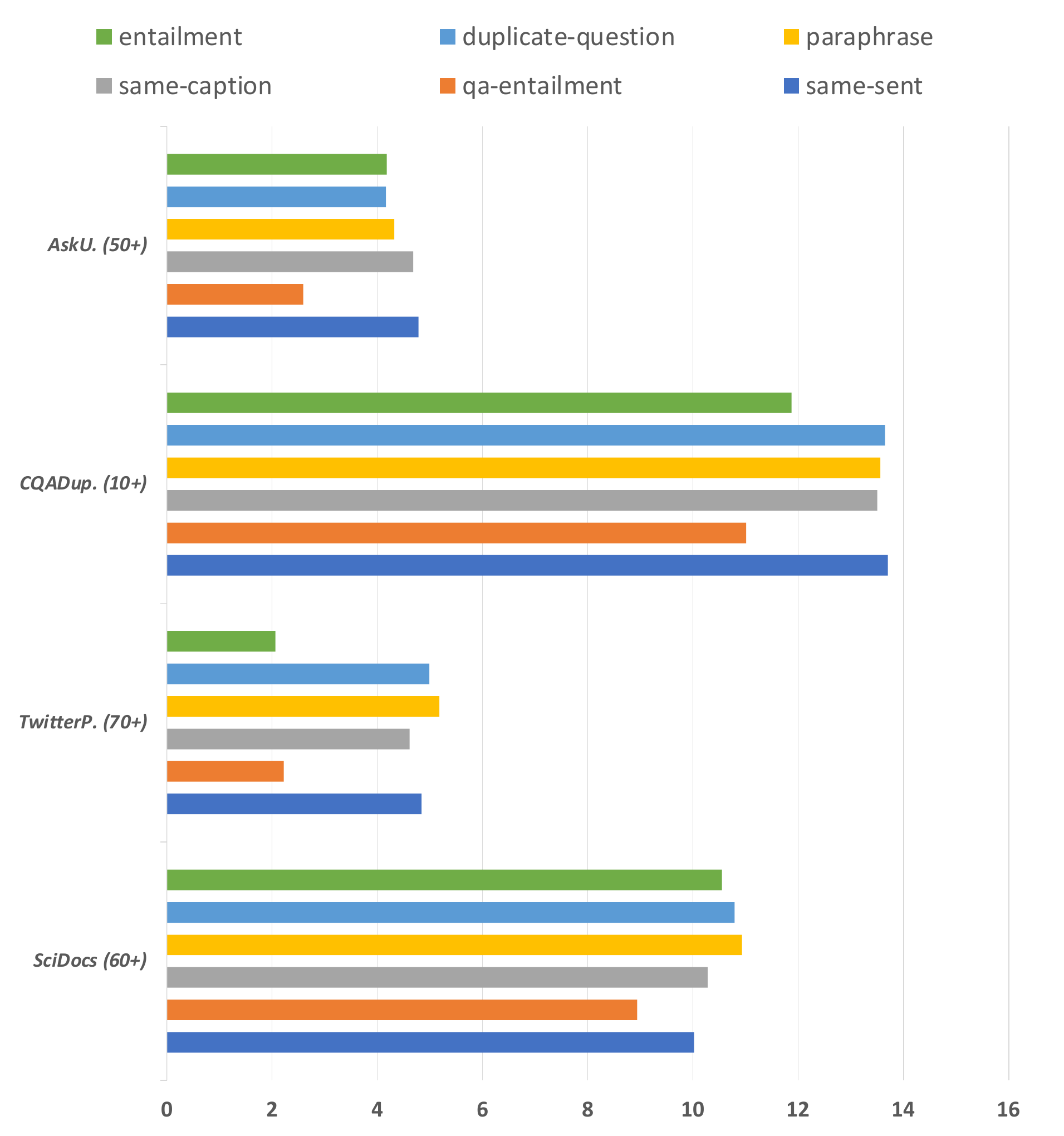}
        \caption{Performance on the USEB benchmark (4 datasets) using different relational similarity scores. The model is BERT\textsubscript{\texttt{base}}. }
        \label{fig:useb-rel-spec}
    \end{figure}

\section{Comparison of Relational Score Selection on USEB Tasks}

    In Section~\ref{sec:useb}, we report the best result on four tasks with the best relational similarity score. We found that different dataset favors different relational similarity score. We look into the detailed performance with all relational similarity scores to further study the effect. The result is shown in Figure~\ref{fig:useb-rel-spec}.

    For AskU. dataset, the unsupervised SimCSE shows the best performance. Similarly, we also found that the unsupervised relation (\emph{same-sent}) demonstrated to be the best relation similarity score to be used in AskU. and CQADup. datasets. TwitterP. dataset is a paraphrase identification task and the similarity score under \emph{paraphrase} relation in RSE shows the best result. In contrast, as no dataset in USEB focuses on question-answer pair retrieval or ranking, the similarity score from \emph{qa-entailment} relation does not lead to a good performance. From the above discussion, we can see that different relational similarity scores will apply to different applications. A wise choice of relational score can lead to better performance, proving that our proposed RSE model can be flexible to various tasks with explicit relation modeling capability.

    \begin{table*}[thb]
        \centering
        \begin{adjustbox}{width=1.0\textwidth,center}
        \begin{tabular}{ l | l | l | c }
        \toprule
        \textbf{Dataset} & \textbf{Relationship} & \textbf{Example Sentence Pair} & \textbf{Tasks} \\ \midrule
        
        \multirow{3}{*}{\begin{tabular}{@{}l@{}} MNLI \& \\ SNLI \end{tabular} } & \multirow{3}{*}{Entailment ($r_1$)} & \begin{tabular}{@{}l@{}} ($s_1$) He mostly hangs out with a group of older, Southern black \\ men, who call him Jumper and Black Cat. \end{tabular} & \multirow{3}{*}{STS, USEB, Transfer} \\
        & & \begin{tabular}{@{}l@{}} ($s_2$) The group of guys he tends to hang out with gave him the \\ nickname Jumper. \end{tabular} & \\ \midrule
        
        \multirow{2}{*}{QQP} & \multirow{2}{*}{Duplicate Question ($r_2$)} & ($s_1$) How can I be a good geologist? & \multirow{2}{*}{STS, USEB} \\
        & & ($s_2$) What should I do to be a great geologist? & \\ \midrule
        
        \multirow{2}{*}{ParaNMT} & \multirow{2}{*}{Paraphrase ($r_3$)} & ($s_1$) he loved that little man , by the way . & \multirow{2}{*}{USEB, Transfer} \\
        & & ($s_2$) he liked the little boy . & \\ \midrule
        
        \multirow{2}{*}{Flicker} & \multirow{2}{*}{Same Caption ($r_4$)} & ($s_1$) Two men in green shirts are standing in a yard. & \multirow{2}{*}{USEB} \\
        & & ($s_2$) Two friends enjoy time spent together. & \\ \midrule
        
        \multirow{2}{*}{QNLI} & \multirow{2}{*}{QA entailment ($r_5$)} & ($s_1$) How many alumni does Olin Business School have worldwide? & \multirow{2}{*}{USEB} \\
        & & ($s_2$) Olin has a network of more than 16,000 alumni worldwide. & \\ \midrule
        
        \multirow{2}{*}{WIKI} & \multirow{2}{*}{Same Sent ($r_6$)} & ($s_1$) A meeting of promoters was also held at Presbyterian Church. & \multirow{2}{*}{USEB} \\
        & & ($s_2$) A meeting of promoters was also held at Presbyterian Church. & \\ \bottomrule
        
        \end{tabular}
        \end{adjustbox} 
        \caption{Example sentence triples for different relations and their applied tasks.}
        \label{tab:7}
    \end{table*}


    


    \begin{table*}[thb]
        \centering
        \begin{adjustbox}{width=1.0\textwidth,center}
        \begin{tabular}{ | c | c | c | c | c | c | c | }
        \toprule
        \textbf{Sentences} & \textbf{Entailment} & \textbf{Duplicate question} & \textbf{Paraphrase} & \textbf{Same caption} & \textbf{QA entailment} & \textbf{Same sent} \\ \midrule \midrule
        \emph{``Where is ACL 2023 hold?''} & \multirow{4}{*}{0.1639} & \multirow{4}{*}{0.3204} & \multirow{4}{*}{0.3037} & \multirow{4}{*}{0.3468} &   \multirow{4}{*}{\textbf{\textcolor{orange}{0.4470}}} & \multirow{4}{*}{0.3078} \\ \cline{1-1}
        \begin{tabular}{@{}c@{}} 
        \emph{``The 61st Annual Meeting of the Association for} \\ 
        \emph{Computational Linguistics (ACL23) will take place} \\ 
        \emph{in Toronto, Canada from July 9th to July 14th, 2023.''} \end{tabular} & & & & & & 
        \\ \midrule \midrule

        \emph{``Is Singapore a city or state?''} & \multirow{3}{*}{0.4220} & \multirow{3}{*}{0.5640} & \multirow{3}{*}{0.5605} & \multirow{3}{*}{0.5774} &   \multirow{3}{*}{\textbf{\textcolor{orange}{0.6599}}} & \multirow{3}{*}{0.5451} \\ \cline{1-1}
        \begin{tabular}{@{}c@{}} 
        \emph{``Singapore is a sunny, tropical island in South-east} \\ 
        \emph{Asia, off the southern tip of the Malay Peninsula.''} \end{tabular} & & & & & & 
        \\ \midrule \midrule

        \emph{``Giraffes can consume 75 pounds of food a day.''} & \multirow{2}{*}{0.8622} & \multirow{2}{*}{\textbf{\textcolor{orange}{0.9596}}} & \multirow{2}{*}{\textbf{\textcolor{orange}{0.9546}}} & \multirow{2}{*}{0.9434} &   \multirow{2}{*}{0.9021} & \multirow{2}{*}{\textbf{\textcolor{orange}{0.9569}}} \\ \cline{1-1}
        \begin{tabular}{@{}l@{}} 
        \emph{``Giraffes can eat up to 75 pounds of food in a day.} \end{tabular} & & & & & & 
        \\ \midrule \midrule

        \begin{tabular}{@{}c@{}} 
        \emph{``Young man breakdancing on a sidewalk in front } \\ 
        \emph{of many people.''} \end{tabular}
        & \multirow{2}{*}{\textbf{\textcolor{orange}{0.6680}}} & \multirow{2}{*}{0.6126} & \multirow{2}{*}{0.6176} & \multirow{2}{*}{0.5606} &   \multirow{2}{*}{0.5062} & \multirow{2}{*}{0.6396} \\ \cline{1-1}
        \begin{tabular}{@{}l@{}} 
        \emph{``A man dancing outside.''} \end{tabular} & & & & & & 
        \\ \bottomrule

        
        
        
        
        
        
        
        \end{tabular}
        \end{adjustbox} 
        \caption{Case study on sentence pairs and the relation scores inferred by our RSE model.}
        \label{tab:case-study}
    \end{table*}

\section{Hyperparameter Settings}

    Table~\ref{tab:7} shows the datasets, relations, examples and their applied tasks in our experiments. We train our model for 3 epochs in all experiments and evaluate every 125, 1000, and 500 steps for best checkpoint selection on STS, USEB, and Transfer tasks, respectively. We select the best model based on its performance on the development set. Due to resource limitations, we perform a small-scale grid search for the model hyperparameter settings. The grid search is on batch size $\in \{128,256,512\}$ and learning rate $\in \{1e-5,3e-5,5e-5\}$. The learning rate for relation embedding is set to $1e-2$ in all experiments because they are randomly initialized parameters without pre-training. All experiments take the representation for \texttt{[CLS]} token as sentence representation. The temperature hyperparameter is set to $0.05$, and the maximum sentence length is set to 32. We did not use warm-up or weight decay in all experiments.

\section{Case Study}

    To better understand the property of RSE, we present the case study of four sentence pairs and their inferred relational scores in Table~\ref{tab:case-study}. The results show that the RSE model can estimate the most likely relationship between sentence pairs. Because the QA entailment relation is very distinct from other relation types, the question-answering pairs are easily distinguished. The QA sentences show the highest score in QA entailment relation, as indicated by the first two cases. Meantime, because of the similar property between relations of \emph{`duplicate question'}, \emph{`paraphrase'} and \emph{`same sent'}, from the third case, we can see highly similar sentence pairs have high relational scores in all three relations. It also aligns with our relation study in Figure~\ref{fig:REL-SIM}. The last case shows two entailment sentence pairs, and we can see that RSE successfully presents the highest score as an entailment relation.







\end{document}